\documentclass[10pt,twocolumn,letterpaper]{article}

\usepackage[pagenumbers]{cvpr} 

\usepackage[pagebackref,breaklinks,colorlinks,allcolors=cvprblue]{hyperref}
\usepackage[utf8]{inputenc} 
\usepackage{CJKutf8} 
\usepackage{mathrsfs}
\usepackage{amsmath}
\usepackage{bm}
\usepackage{multirow}
\usepackage{bbding}
\usepackage{tcolorbox}
\usepackage{colortbl}

\definecolor{rred}{RGB}{245, 152, 153}
\definecolor{oorange}{RGB}{253, 205, 154}
\definecolor{cvprblue}{rgb}{0.21,0.49,0.74}

\usepackage{hyperref}
\hypersetup{
colorlinks,
linkcolor={red},
urlcolor={purple}
}

\title{MatAnyone 2: Scaling Video Matting via a Learned Quality Evaluator}

\author{Peiqing Yang$^{1}$ \quad Shangchen Zhou$^{1}$$^{\dag}$ \quad Kai Hao$^{1}$ \quad Qingyi Tao$^{2}$\\
$^{1}$S-Lab, Nanyang Technological University \quad $^{2}$SenseTime Research, Singapore\\
{\tt\small \url{https://pq-yang.github.io/projects/MatAnyone2/}}
}

\begin{document}
\begin{CJK}{UTF8}{gbsn}

\twocolumn[{%
\renewcommand\twocolumn[1][]{#1}%
\maketitle
\vspace{-10mm}
\begin{center}
    \centering
    \includegraphics[width=\linewidth]{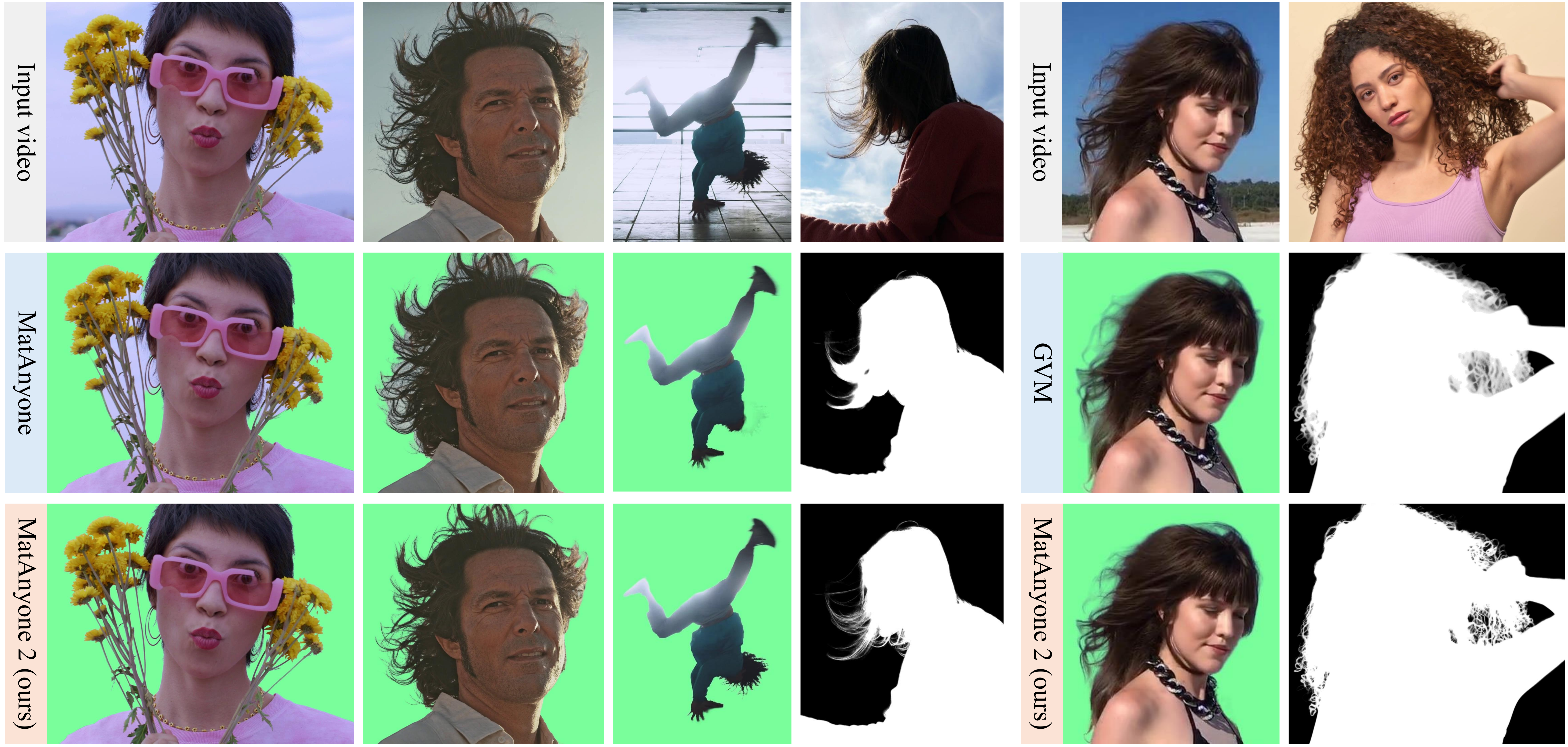}
    \vspace{-5.5mm}
    \captionof{figure}{
    (a) Our MatAnyone 2 significantly outperforms MatAnyone~\cite{yang2025matanyone} in preserving fine details and avoiding segmentation-like boundaries, while also showing enhanced robustness under challenging lighting conditions, \eg, backlit scenes.
    (b) As a diffusion-based video matting method, GVM~\cite{ge2025gvm} often produces blurry alpha mattes with unnatural transitions along object boundaries, \eg, hair strands. In contrast, MatAnyone 2 generates clear, high-quality alpha mattes with natural boundary details.
    \textbf{(Zoom-in for best view)}
    } \vspace{1mm}
    \label{fig:teaser}
\end{center}
}]

\def\thefootnote{\dag}\footnotetext{Project lead.\vspace{-1mm}}

\begin{abstract}
Video matting remains limited by the scale and realism of existing datasets. While leveraging segmentation data can enhance semantic stability, the lack of effective boundary supervision often leads to segmentation-like mattes lacking fine details. 
To this end, we introduce a learned Matting Quality Evaluator (MQE) that assesses semantic and boundary quality of alpha mattes without ground truth. It produces a pixel-wise evaluation map that identifies reliable and erroneous regions, enabling fine-grained quality assessment.
The MQE scales up video matting in two ways: (1) as an online matting-quality feedback during training to suppress erroneous regions, providing comprehensive supervision, and (2) as an offline selection module for data curation, improving annotation quality by combining the strengths of leading video and image matting models. This process allows us to build a large-scale real-world video matting dataset, VMReal, containing 28K clips and 2.4M frames.
To handle large appearance variations in long videos, we introduce a reference-frame training strategy that incorporates long-range frames beyond the local window for effective training.
Our MatAnyone 2 achieves state-of-the-art performance on both synthetic and real-world benchmarks, surpassing prior methods across all metrics.
\end{abstract}    
\vspace{-2mm}
\section{Introduction}
\label{sec:intro}

Video matting (VM) has gained increasing attention for its broad applications in visual effects and video editing. Despite recent advances~\cite{yang2025matanyone,huynh2024maggie,ge2025gvm,lin2023adam,huang2023ftp,lin2022rvm}, existing methods still suffer from blurry boundaries, missing regions, and unstable tracking. A key underlying reason is the limited scale, quality, and realism of existing VM datasets.
The recent VM800 dataset~\cite{yang2025matanyone} provides the largest collection of video foregrounds, but it contains only 826 sequences ($\sim$ 1/60 the size of the video object segmentation (VOS) dataset used in SAM 2~\cite{ravi2024sam2}).
To enrich training samples, existing pipelines typically composite foregrounds onto random backgrounds via RGBA blending. However, such synthetic data often exhibit inconsistent lighting and unnatural boundaries, leading to a domain gap from real videos that limits model generalization in challenging cases.

To alleviate data limitations, recent video matting methods have explored leveraging segmentation priors from pretrained models and large-scale VOS data.
(1) \textit{Pretraining.} Some works employ segmentation models~\cite{huynh2024maggie} or data~\cite{ge2025gvm,lin2023adam,zhang2025oavm} for pretraining, followed by finetuning on matting datasets. However, the limited availability of high-quality matting data often causes the pretrained segmentation capability to degrade during finetuning.
(2) \textit{Joint-training.} Some studies~\cite{lin2022rvm,huang2023ftp,yang2025matanyone} instead combine segmentation and matting supervision within a unified framework, enabling the model to learn both tasks simultaneously.
However, segmentation data can only provide reliable supervision for \textit{non-boundary regions} (where alpha values are 0 or 1). For \textit{boundary regions}, MatAnyone~\cite{yang2025matanyone} applies an unsupervised loss~\cite{liu2024ddc} that relies on a strong assumption, leading to weak boundary supervision. While this enhances semantic and tracking accuracy, it often causes the predicted mattes degrade into segmentation-like masks, especially around boundaries (see Fig.~\ref{fig:teaser}).
These methods seek an optimal trade-off between matting and segmentation through balanced data and loss, yet simultaneously improving semantic accuracy and boundary details remains highly challenging.

To this end, we introduce a novel \textit{Matting Quality Evaluator (MQE)} that mitigates the long-standing shortage of real VM data and effectively harnesses segmentation information.
This evaluator operates in two modes: (i) an \textit{online matting-quality guidance} during training, and (ii) an \textit{offline pixel-wise selection module} for data curation, which together scale up video matting training.
Given an input frame, its segmentation mask, and the predicted alpha matte, the evaluator produces a binary evaluation map that identifies ``\textit{reliable}’’ vs. ``\textit{erroneous}’’ regions.
It implicitly inherits the binary labels of non-boundary regions from the input segmentation masks, while assessing fine-detail matting quality in boundary regions leveraging a DINOv3-based encoder~\cite{simeoni2025dinov3}. This enables the learned MQE to accurately identify inaccurate or low-quality alpha predictions in a pixel-wise manner, thereby providing a comprehensive assessment of both semantic accuracy and detail fidelity (see Fig.~\ref{fig:qe_eval}), \textit{without} requiring ground-truth mattes.

\noindent \textit{Online Guidance} -
MQE serves as an online guidance signal that provides dynamic supervision through its on-the-fly quality assessment. Specifically, its binary evaluation map acts as a penalty map that encourages the network to correct errors in the \textit{erroneous} regions. Operating without ground-truth mattes (requiring only segmentation masks), MQE offers effective boundary supervision for rich segmentation data, thereby enabling large-scale training for matting.

\noindent \textit{Offline Selection} -
MQE also serves as an offline selection module for \textit{scalable} curation of a real-world video matting dataset.
Manual annotation of video alpha mattes is extremely challenging, if not impossible at scale, leaving such a dataset long absent and hindering progress in this field.
To overcome this, we design an automated dual-branch annotation pipeline, where MQE acts as a quality arbiter to assess and fuse predictions from two complementary methods: state-of-the-art video and image matting models, which excel in semantic stability and boundary detail, respectively. By selecting reliable regions from each, the MQE combines their strengths, yielding annotation with both stable semantics and high-quality details.
This pipeline constructs \textit{VMReal}, a \textit{large-scale} and \textit{real-world} video matting dataset with about 28K clips ($\sim 35\times$ larger than previous dataset~\cite{yang2025matanyone}) and 2.4M annotated frames covering diverse scenes, lighting conditions, and motion patterns.

Moreover, in long videos where subject appearance varies largely over time, propagation-based methods like MatAnyone~\cite{yang2025matanyone} may fail to matte out new human or clothing parts unseen in earlier frames. Training on short sequences cannot model such variations, while long sequences lead to excessive memory overhead. To address this, we introduce a reference-frame strategy that incorporates long-range frames beyond the local training window, enabling the model to handle large appearance changes across long videos without large memory costs.

In summary, our main contributions are as follows:

\begin{itemize}
    \item We present a learned MQE that performs semantic and detail assessment of predicted mattes without ground truth. It functions as both an online guidance signal for training and an offline selection module for data curation, effectively scaling up video matting.
    \item Leveraging the MQE, we construct \textit{VMReal}, a large-scale, scene-diverse, and real-world video matting dataset, with about 28K clips and 2.4M annotated frames.
    \item We introduce a reference-frame training strategy that expands the temporal context, allowing our method to handle large appearance variations in long videos efficiently.
\end{itemize}
Together, these contributions lead to MatAnyone 2, which achieves superior results on both synthetic and real-world benchmarks, surpassing existing methods across all metrics.

\section{Related Work}
\label{sec:relatedwork}

\begin{figure*}[t]
\begin{center}
    \vspace{-3mm}
    \includegraphics[width=\linewidth]{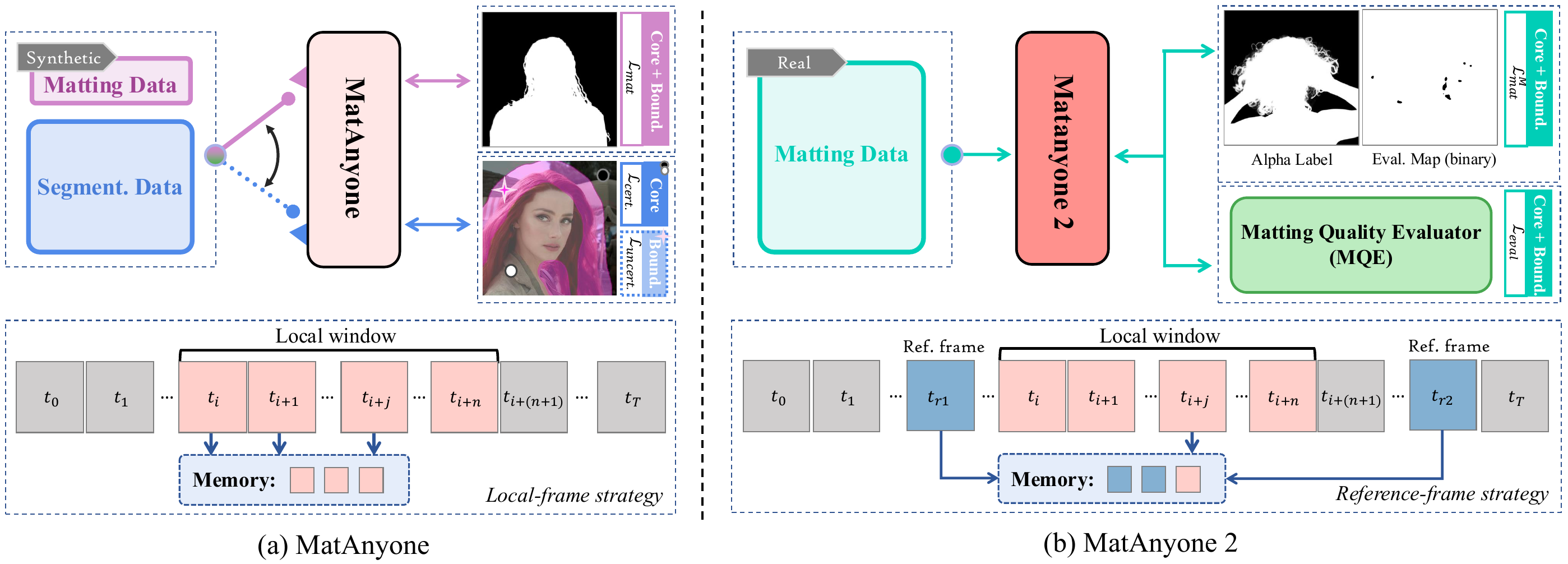}
    \vspace{-6mm}
    \caption{
    Comparison of MatAnyone and MatAnyone 2.
    (a) MatAnyone~\cite{yang2025matanyone} compensates for limited VM data by leveraging segmentation supervision, which provides reliable core-region guidance ($\mathcal{L}_{cert.}$) but weak boundary supervision ($\mathcal{L}_{uncert.}$), leading to segmentation-like mattes. Its memory is restricted to the local training window, limiting its ability to model large appearance changes.
    (b) MatAnyone~2 is trained with our real-world VMReal dataset, where each alpha label is paired with a binary evaluation map. This enables masked matting loss $\mathcal{L}_{mat}^M$ on reliable pixels, while the learned MQE supplies $\mathcal{L}_{eval}$ to supervise both core and boundary regions. For large variations, a reference-frame strategy adds long-range temporal cues, improving robustness to large appearance changes without extra memory cost.
    }
    \vspace{-6mm}
\label{fig:matanyone1vs2}
\end{center}
\end{figure*}

\noindent \textbf{Video Matting.}
In the absence of auxiliary inputs~\cite{zhang2019late,qiao2020attention,lin2022rvm,ke2022MODNet,li2024vmformer,zhu2017fast,qin2025bivm}, VM often becomes object-specific, with human video matting~\cite{shen2016deep,zhu2017fast,ke2022MODNet,li2024vmformer,ge2025gvm} being the most widely studied due to its strong real-world applicability.
Nevertheless, these approaches tend to fail in challenging backgrounds when other visually similar objects appear in the scene.
Recent progress in promptable segmentation~\cite{kirillov2023sam,ravi2024sam2,zou2024seem,zhou2023edgesam} has enabled human masks to be obtained with minimal effort, inspiring a series of mask-guided image~\cite{yao2024matteanything,yao2024vitmatte,cai2022transmatting,li2024matting,github:mattepro} and video matting methods~\cite{huang2023ftp,lin2023adam,huynh2024maggie,li2024vim,zhang2025oavm,yang2025matanyone} that incorporate segmentation priors for improved robustness across scenes.
With the emergence of video diffusion models~\cite{blattmann2023svd,yang2024cogvideox}, several recent studies have replaced traditional CNN backbones with diffusion networks~\cite{ge2025gvm,yang2025vrmdiff}, reframing video matting as a generation problem. For example, GVM~\cite{ge2025gvm} leverages rich priors from a pretrained video diffusion model~\cite{blattmann2023svd} by conditioning on the input video.
Although the diffusion prior provides robustness, GVM still yields blurry alpha mattes, revealing its limitation imposed by suboptimal training data.

\noindent \textbf{Video Matting and Segmentation Datasets.}
Existing video matting datasets~\cite{lin2021bgm,yang2025matanyone,ge2025gvm,huynh2024maggie,yang2025vrmdiff,wang2021crgnn,zhang2021vm108} suffer from limited scale, quality, and realism, which significantly constrain model performance and generalization.
Although recent efforts such as VM800~\cite{yang2025matanyone} have expanded data scale and quality compared to VideoMatte240K~\cite{lin2021bgm}, and GVM~\cite{ge2025gvm} constructs high-fidelity 4K rendered clips, these datasets remain synthetic and are still much smaller than large-scale video object segmentation (VOS) datasets~\cite{ravi2024sam2,yang2019vos}.
For instance, the SA-V dataset used by SAM~2~\cite{ravi2024sam2} is roughly 60 times larger than VM800 and contains diverse real-world videos with dense annotations.
While real-world image matting datasets such as P3M-10k~\cite{li2021p3m}, D646~\cite{qiao2020attention}, and AIM~\cite{AIM} are available, they are limited in scale and, when used for training alone, can harm the temporal consistency of video matting models.
In this work, we construct VMReal, a large-scale real-world video matting dataset, to address the data deficiency in the video matting field.

\noindent \textbf{Quality Assessment for Data Curation.}
Due to the high cost of manual annotation, large-scale datasets often rely on model-assisted or fully automated labeling pipelines. Recent works in image generation~\cite{wallace2024diffusion, na2025boost, liu2025improving, wu2025qwen} and restoration~\cite{wu2025dp,cai2025dspo} employ image quality assessment (IQA) models to construct better–worse data pairs within the Direct Preference Optimization (DPO) framework, guiding models toward perceptually better results.
Also, automatic quality evaluation has been explored in low-level vision tasks~\cite{wu2025dp,cai2025dspo,yang2024depthv2,li2025rorem}. For instance, Depth Anything V2~\cite{yang2024depthv2} integrates multi-model voting with selective human verification to ensure annotation reliability, and RORem~\cite{li2025rorem} leverages human feedback to train a discriminator for automatic high-quality sample selection. These methods assess quality at the image level, whereas video matting demands pixel-level and temporally consistent evaluation, motivating our design of a learned MQE for constructing a large-scale video matting dataset.

\section{Methodology}
\label{sec:method}

In this study, our goal is to scale up video matting with both \textit{high semantic accuracy} and \textit{fine boundary fidelity}.
Previous work, MatAnyone~\cite{yang2025matanyone}, improves video matting by jointly training with segmentation data (Fig.~\ref{fig:matanyone1vs2}(a)) through an improved supervision structure and loss design. However, segmentation data can only provide reliable supervision for non-boundary regions, while boundary regions are weakly constrained by an unsupervised loss. Although this improves semantic accuracy, it inevitably degrades boundary fidelity, resulting in segmentation-like mattes (see Fig.~\ref{fig:teaser}).
To this end, we propose MatAnyone 2 (Fig.~\ref{fig:matanyone1vs2}(b)), featuring a novel \textit{MQE}.
The MQE provides effective supervision for both non-boundary and boundary regions, without requiring matting ground truth.
Moreover, MQE facilitates the construction of a large-scale real-world video matting dataset, allowing VM models to be effectively trained on high-quality matting data. Thanks to the improved supervision and training data brought by MQE, MatAnyone 2 achieves higher semantic accuracy while preserving fine boundary details.
In the following sections, we first introduce the training of the MQE model (Sec.~\ref{subsec:qe_model}), followed by its two roles for scaling up video matting: (i) an \textit{online matting-quality guidance} during training (Sec.~\ref{subsec:online}), and (ii) an \textit{offline pixel-wise selection module} for data curation (Sec.~\ref{subsec:offline}). Finally, we present a reference-frame training strategy that incorporates long-range frames to efficiently handle large appearance variations in long videos (Sec.~\ref{subsec:ref_frame}).

\begin{figure}[t]
\begin{center}
    \includegraphics[width=\linewidth]{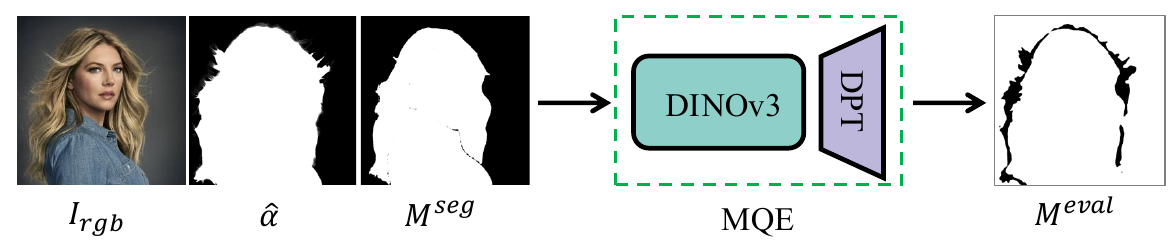}
    \vspace{-6mm}
    \caption{
    Given an input tuple of video frame $I_{rgb}$, predicted alpha $\hat{\alpha}$, and segmentation mask $M^{seg}$, MQE produces a binary evaluation map $M^{eval}$ at the pixel-wise level (1: \textit{reliable}, 0: \textit{erroneous}).
    }
    \vspace{-8mm}
\label{fig:qe}
\end{center}
\end{figure}

\subsection{Quality Evaluation Model}
\label{subsec:qe_model}
We aim to develop a reliable and pixel-wise MQE model that has the potential to provide effective matting-quality feedback during training while also assisting in data curation. As shown in Fig.~\ref{fig:qe}, our MQE network receives a tuple input $I_{in}^{eval} := <I_{rgb}, \hat{\alpha}, M^{seg}>$, consisting of the video frame, predicted alpha matte, and segmentation mask, and produces a binary evaluation map $M^{eval} \in \{0,1\}^{H \times W}$, where 1 and 0 indicate \textit{reliable} and \textit{erroneous} regions, respectively.

\begin{figure}[t]
\begin{center}
    \includegraphics[width=0.98\linewidth]{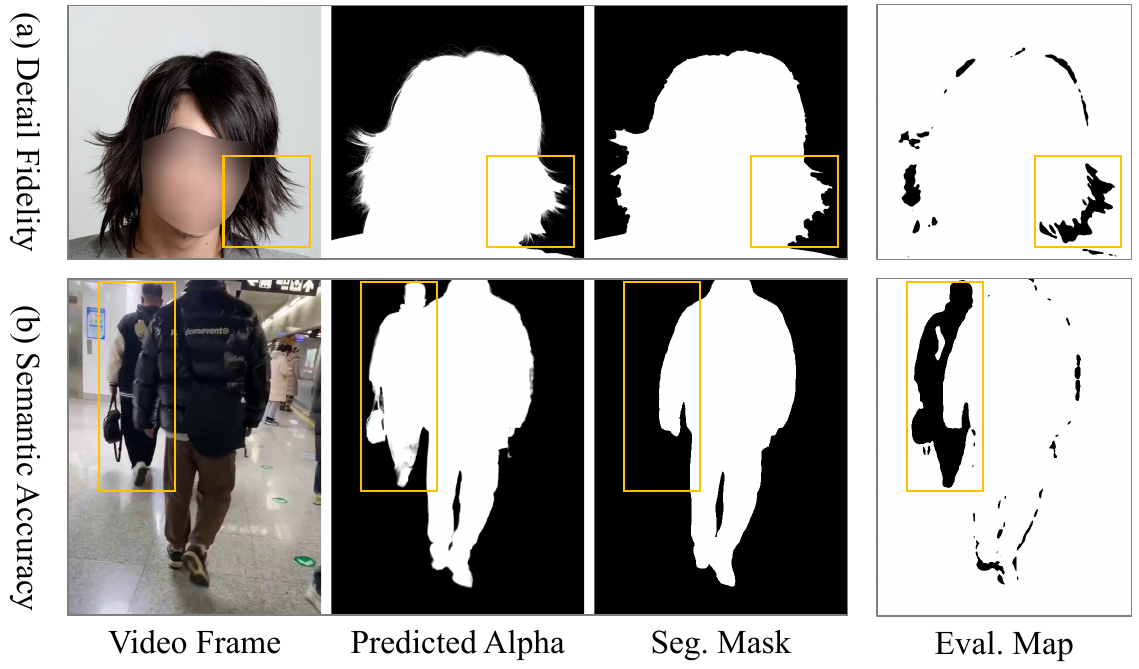}
    \vspace{-1mm}
    \caption{
    Our learned Matting Quality Evaluation (MQE) model accurately identifies: (a) low-quality matting details along the boundary, and (b) semantically wrong regions at core areas, in a pixel-wise manner \textit{without} requiring ground-truth mattes.
    }
    \vspace{-8mm}
\label{fig:qe_eval}
\end{center}
\end{figure}

Providing the segmentation mask as an auxiliary input is crucial for achieving reliable matting quality evaluation.
As shown in Fig.~\ref{fig:qe_eval}(b), for \textit{non-boundary regions}, the learned MQE tends to inherit semantic cues from the segmentation mask, allowing it to easily identify foreground and background areas. For \textit{boundary regions}, the MQE leverages the segmentation map to guide its attention toward fine-grained matting quality, such as hair mattes in Fig.~\ref{fig:qe_eval}(a). Moreover, as our MQE employs DINOv3~\cite{simeoni2025dinov3} as the encoder, it benefits from the rich and high-quality features, enabling it to effectively perceive and identify inaccurate or low-quality alpha predictions.

\noindent \textbf{Training and Data for MQE.}
We formulate MQE learning as a binary segmentation task. To train such a model, we require a binary ground-truth map $M^{eval}_{gt} \in \{0,1\}^{H \times W}$ for evaluating an alpha prediction $\hat{\alpha}$, which does not naturally exist. Therefore, we construct training pairs using the existing image matting dataset P3M-10k~\cite{li2021p3m}, which provides human-annotated alpha mattes $\alpha_{gt}$ corresponding to images $I_{rgb}$.
To evaluate $\hat{\alpha}$ against $\alpha_{gt}$ and generate $M^{eval}_{gt}$, we design a discrepancy measure $\mathcal{D}(\cdot)$ that computes a weighted combination of standard video matting metrics (MAD and Grad) within local patches. This measure effectively captures both semantic- and detail-level deviations. Specifically, the ground-truth evaluation map is defined as $M^{eval}_{gt} = \mathbb{I}\big(\mathcal{D}(\alpha_{gt}, \hat{\alpha}) < \delta\big)$, where $\mathbb{I}(\cdot)$ denotes the indicator function and $\delta$ is a threshold.
Since erroneous regions primarily occur around boundaries, class 1 (reliable regions) dominate class 0 (erroneous regions) in the training data, resulting in class imbalance. To mitigate this, we adopt the focal loss~\cite{lin2017focal} to emphasize hard erroneous regions during training.

%
\subsection{Online Matting-quality Guidance}
\label{subsec:online}
Thanks to the pixel-wise quality evaluation provided by the learned MQE model, we leverage it as a matting-quality feedback signal during training, where its on-the-fly assessment offers dynamic online supervision, as shown in Fig.~\ref{fig:matanyone1vs2}(b). 
For each input frame, the MQE first produces a probability map $P^{(0)}_{eval}$ that indicates the likelihood of error for its alpha prediction in a pixel-wise manner, without requiring ground-truth alpha mattes. This map can serve as a penalty mask to guide the network in suppressing erroneous regions. The online guidance loss is defined as:
\begin{equation}
\mathcal{L}_{eval} = \| P^{(0)}_{eval} \|_1,
\end{equation}
which encourages lower error probabilities and thus more accurate alpha predictions.
Compared to the weak unsupervised loss used in MatAnyone~\cite{yang2025matanyone}, our MQE-based guidance loss $\mathcal{L}_{eval}$ provides a more effective and stable learning signal for boundary regions (see Table~\ref{tab:ablation}(b)), enabling large-scale video matting training with unlabeled videos.

\begin{figure*}[t!]
\begin{center}
    \vspace{-2mm}
    \includegraphics[width=\linewidth]{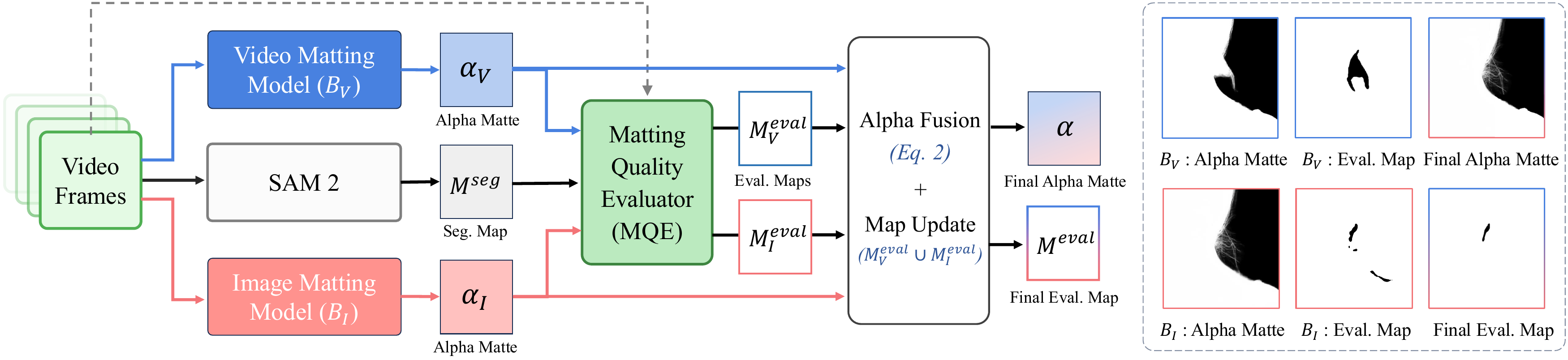}
    \vspace{-6mm}
    \caption{
    Our automated dual-branch annotation pipeline.
    This pipeline enables large-scale construction of real-world VM datasets, resulting in our VMReal dataset.
    We combine two complementary annotation branches: the temporally stable $B_V$ branch provides the base annotation, while the detail-preserving $B_I$ branch offers fine boundary details.
    Pixels where $B_V$ fails but $B_I$ succeeds are selectively integrated to produce the fused alpha and its evaluation map.
    The example on the right shows that the fused annotation preserves semantic stability as well as rich boundary details, making it suitable for VM training.
    }
    \vspace{-5mm}
\label{fig:data_pipeline}
\end{center}
\end{figure*}

\subsection{Offline Pixel-wise Selection}
\label{subsec:offline}
Beyond using MQE as matting-quality guidance during training, we further employ it to directly construct a large-scale, high-quality, real-world video matting dataset with detailed alpha annotations.
Manual annotation of video alpha mattes \textit{at scale} is prohibitively expensive, if not impossible, making such a dataset long unattainable and hindering progress in this field. 
To this end, we design an automated dual-branch annotation pipeline, as shown in Fig.~\ref{fig:data_pipeline}, where MQE acts as a quality arbiter to evaluate and fuse predictions from two complementary branches, employing state-of-the-art video and image matting models.

\noindent \textbf{Dual-branch Annotation.}
The first branch ($B_V$) employs a video matting model, \eg, MatAnyone~\cite{yang2025matanyone}, which ensures stable temporal consistency and robust overall structure but sometimes lacks fine details around object boundaries.
In contrast, the second branch ($B_I$) adopts an image matting model, \eg, MattePro~\cite{github:mattepro}, guided by per-frame segmentation masks from SAM 2~\cite{ravi2024sam2}. Although it struggles to maintain temporal stability and semantic consistency, it produces sharper boundary details that effectively complement the video-matting branch.
For each video sequence, these two branches generate complementary annotations: $B_V$ excels in semantic stability, while $B_I$ provides high-quality boundary details.

\noindent \textbf{MQE as Quality Arbiter.}
The learned MQE acts as a quality arbiter that identifies reliable regions from each branch and fuses their strengths in a pixel-wise manner.
As shown in Fig.~\ref{fig:data_pipeline}, given video-based data, the two branches $B_V$ and $B_I$ respectively produce alpha mattes $\alpha_{V}$ and $\alpha_{I}$. The MQE then performs quality evaluation on both branches, generating corresponding evaluation maps $M^{eval}_{V}$ and $M^{eval}_{I}$.
We take the stable and temporally consistent predictions from $B_V$ as the base, while MQE selectively integrates boundary details from $B_I$ in regions where $B_V$ is unreliable ($M^{eval}_{V}=0$) but $B_I$ is reliable ($M^{eval}_{I}=1$). 
We denote such regions with a fusion mask, \ie, $M^{fuse}=M^{eval}_{I} \odot (1 - M^{eval}_{V})$, which is used to blend the two branches.
To avoid boundary artifacts, we smooth the mask $M^{fuse}$ with a Gaussian blur, resulting in seamless and soft blending between the two branches.
The final fused alpha matte is obtained by:
\begin{equation}
\alpha = \alpha_{V} \odot (1-M^{fuse}) + \alpha_{I} \odot M^{fuse},
\end{equation}
which preserves both semantic stability of $B_V$ and fine boundary details from $B_I$.
Meanwhile, the fused evaluation map is updated as $M^{eval}=M^{eval}_{V}\cup M^{eval}_{I}$.

\noindent \textbf{VMReal Dataset.}
Leveraging the proposed MQE-based dual-branch annotation pipeline, we construct \textit{VMReal}, a large-scale, scene-diverse, and real-world video matting dataset with high-quality alpha annotations, comprising about 28K clips ($\sim 35\times$ larger than the previous dataset~\cite{yang2025matanyone}) and 2.4M frames.
Among them, a high-quality subset of 4.5K clips is collected from video footage websites and YouTube, featuring 1080p resolution with abundant fine-grained hair details.
The remaining videos are derived from the SA-V dataset~\cite{ravi2024sam2} (we filter out a human-centric subset), typically at 720p resolution, offering rich diversity in scenes, motions, and lighting conditions.

\noindent \textbf{Train with VMReal Dataset.}
With the scalable automatic annotation pipeline, we can convert large amounts of unlabeled videos into matting data for training.
Thus, we no longer perform joint training with both matting and segmentation data, effectively avoiding the segmentation-like results and loss of fine boundary details often caused by segmentation supervision.
Instead, all samples are unified through our annotation pipeline into triplets of $\langle I_{rgb}, \alpha, M^{eval} \rangle$, where $M^{eval}$ denotes the pixel-wise reliability map.
During training, the matting loss $\mathcal{L}_{mat}^M$ is computed only over reliable regions where $M^{eval}=1$, ensuring that supervision focuses on high-confidence areas while ignoring uncertain or low-quality regions.
In this way, our unified training scheme as shown in Fig.~\ref{fig:matanyone1vs2}(b) fully leverages diverse heterogeneous data sources and effectively scales up video matting training.

\begin{table*}[t!]
\begin{center}
\setlength{\fboxsep}{2.8pt}
\caption{
    Quantitative comparisons on different video matting benchmarks from diverse sources. The best and second-best performances are marked in \colorbox{rred}{\underline{red}} and \colorbox{oorange}{orange}, respectively.
    $*$ indicates that GVM~\cite{ge2025gvm} is a diffusion-based video matting method that leverages rich Stable Video Diffusion~\cite{blattmann2023svd} priors, while our method is purely CNN-based.
    $\dag$ indicates that MaGGIe~\cite{huynh2024maggie} requires the instance mask as guidance for each frame, while our method only requires it in the first frame.
    }
\label{tab:comparison_syn}
\vspace{-1mm}
\renewcommand{\arraystretch}{1.15}
\renewcommand{\tabcolsep}{3.0mm}
\scalebox{0.78} {
\begin{tabular}{lccccccccc}
\hline
\multicolumn{1}{l|}{\multirow{2}{*}{Metrics}} &
  \multicolumn{4}{c|}{\textbf{Auxiliary-free (AF) Methods}} &
  \multicolumn{5}{c}{\textbf{Mask-guided Methods}} \\ \cline{2-10} 
\multicolumn{1}{l|}{} &
  MODNet~\cite{ke2022MODNet} &
  RVM~\cite{lin2022rvm} &
  RVM-Large~\cite{lin2022rvm} &
  \multicolumn{1}{c|}{GVM$^{*}$~\cite{ge2025gvm}} &
  AdaM~\cite{lin2023adam} &
  FTP-VM~\cite{huang2023cvpr} &
  MaGGIe$^{\dag}$~\cite{huynh2024maggie} &
  MatAnyone~\cite{yang2025matanyone} &
  \textbf{Ours} \\ \hline \hline
\multicolumn{9}{l}{\textbf{\textit{VideoMatte}} ($512 \times 288$)} \\ \hline
\multicolumn{1}{l|}{MAD$\downarrow$} &
  \multicolumn{1}{c}{9.41} &
  \multicolumn{1}{c}{6.08} &
  \multicolumn{1}{c}{5.32} &
  \multicolumn{1}{c|}{6.39} &
  \multicolumn{1}{c}{5.30} &
  \multicolumn{1}{c}{6.13} &
  \multicolumn{1}{c}{5.49} &
  \multicolumn{1}{c}{\colorbox{oorange}{5.15}} &
  \multicolumn{1}{c}{\colorbox{rred}{\underline{4.73}}} \\
\multicolumn{1}{l|}{MSE$\downarrow$} &
  4.30 &
  1.47 &
  \multicolumn{1}{c}{0.62} &
  \multicolumn{1}{c|}{1.82} &
  0.78 &
  1.31 &
  \colorbox{oorange}{0.60} &
  0.93 &
  \colorbox{rred}{\underline{0.55}} \\
\multicolumn{1}{c|}{Grad$\downarrow$} &
  1.89 &
  0.88 &
  \multicolumn{1}{c}{0.59} &
  \multicolumn{1}{c|}{0.72} &
  0.72 &
  1.14 &
  \colorbox{oorange}{0.57} &
  0.67 &
  \colorbox{rred}{\underline{0.51}} \\
\multicolumn{1}{l|}{dtSSD$\downarrow$} &
  \multicolumn{1}{c}{2.23} &
  \multicolumn{1}{c}{1.36} &
  \multicolumn{1}{c}{1.24} &
  \multicolumn{1}{c|}{1.27} &
  \multicolumn{1}{c}{1.33} &
  \multicolumn{1}{c}{1.60} &
  \multicolumn{1}{c}{1.39} &
  \multicolumn{1}{c}{\colorbox{oorange}{1.18}} &
  \multicolumn{1}{c}{\colorbox{rred}{\underline{1.12}}} \\
\multicolumn{1}{l|}{Conn$\downarrow$} &
  0.81 &
  0.41 &
  \multicolumn{1}{c}{{0.30}} &
  \multicolumn{1}{c|}{0.43} &
  {0.30} &
  0.41 &
  0.31 &
  \colorbox{oorange}{0.26} &
  \colorbox{rred}{\underline{0.20}} \\ \hline
\multicolumn{9}{l}{\textbf{\textit{VideoMatte}} ($1920 \times 1080$)} \\ \hline
\multicolumn{1}{l|}{MAD$\downarrow$} &
  11.13 &
  6.57 &
  5.81 &
  \multicolumn{1}{c|}{6.33} &
  {4.42} &
  8.00 &
  {4.42} &
  \colorbox{oorange}{4.24} &
  \colorbox{rred}{\underline{4.10}} \\
\multicolumn{1}{l|}{MSE$\downarrow$} &
  \multicolumn{1}{c}{5.54} &
  \multicolumn{1}{c}{1.93} &
  \multicolumn{1}{c}{0.97} &
  \multicolumn{1}{c|}{2.08} &
  \multicolumn{1}{c}{{0.39}} &
  \multicolumn{1}{c}{3.24} &
  \multicolumn{1}{c}{0.40} &
  \multicolumn{1}{c}{\colorbox{oorange}{0.33}} &
  \multicolumn{1}{c}{\colorbox{rred}{\underline{0.28}}} \\
\multicolumn{1}{l|}{Grad$\downarrow$} &
  \multicolumn{1}{c}{15.30} &
  \multicolumn{1}{c}{10.55} &
  \multicolumn{1}{c}{9.65} &
  \multicolumn{1}{c|}{8.04} &
  \multicolumn{1}{c}{5.12} &
  \multicolumn{1}{c}{23.75} &
  \multicolumn{1}{c}{{4.03}} &
  \multicolumn{1}{c}{\colorbox{oorange}{4.00}} &
  \multicolumn{1}{c}{\colorbox{rred}{\underline{3.45}}} \\
\multicolumn{1}{l|}{dtSSD$\downarrow$} &
  3.08 &
  1.90 &
  \multicolumn{1}{c}{1.78} &
  \multicolumn{1}{c|}{1.59} &
  1.39 &
  2.37 &
 {1.31} &
  \colorbox{oorange}{1.19} &
  \colorbox{rred}{\underline{1.15}} \\ \hline \hline
  \multicolumn{9}{l}{\textbf{\textit{YoutubeMatte}} ($512 \times 288$)} \\ \hline
\multicolumn{1}{l|}{MAD$\downarrow$} &
  \multicolumn{1}{c}{19.37} &
  \multicolumn{1}{c}{4.08} &
  \multicolumn{1}{c}{3.36} &
  \multicolumn{1}{c|}{3.30} &
  \multicolumn{1}{c}{-} &
  \multicolumn{1}{c}{{3.08}} &
  \multicolumn{1}{c}{3.54} &
  \multicolumn{1}{c}{\colorbox{oorange}{2.72}} &
  \multicolumn{1}{c}{\colorbox{rred}{\underline{2.30}}} \\
\multicolumn{1}{l|}{MSE$\downarrow$} &
  16.21 &
  1.97 &
  \multicolumn{1}{c}{{1.04}} &
  \multicolumn{1}{c|}{1.52} &
  - &
  1.29 &
  1.23 &
  \colorbox{oorange}{1.01} &
  \colorbox{rred}{\underline{0.78}} \\
\multicolumn{1}{c|}{Grad$\downarrow$} &
  2.05 &
  1.34 &
  \multicolumn{1}{c}{1.03} &
  \multicolumn{1}{c|}{\colorbox{oorange}{0.79}} &
  - &
  1.16 &
  1.10 &
  0.97 &
  \colorbox{rred}{\underline{0.78}} \\
\multicolumn{1}{l|}{dtSSD$\downarrow$} &
  \multicolumn{1}{c}{2.79} &
  \multicolumn{1}{c}{1.81} &
  \multicolumn{1}{c}{{1.62}} &
  \multicolumn{1}{c|}{1.52} &
  \multicolumn{1}{c}{-} &
  \multicolumn{1}{c}{1.83} &
  \multicolumn{1}{c}{1.88} &
  \multicolumn{1}{c}{\colorbox{oorange}{1.60}} &
  \multicolumn{1}{c}{\colorbox{rred}{\underline{1.45}}} \\
\multicolumn{1}{l|}{Conn$\downarrow$} &
  2.68 &
  0.60 &
  \multicolumn{1}{c}{0.50} &
  \multicolumn{1}{c|}{0.44} &
  - &
 {0.41} &
  0.49 &
  \colorbox{oorange}{0.39} &
  \colorbox{rred}{\underline{0.32}} \\ \hline
\multicolumn{9}{l}{\textbf{\textit{YoutubeMatte}} ($1920 \times 1080$)} \\ \hline
\multicolumn{1}{l|}{MAD$\downarrow$} &
  15.29 &
  4.37 &
  \multicolumn{1}{c}{3.58} &
  \multicolumn{1}{c|}{2.68} &
  - &
  6.49 &
  {2.37} &
  \colorbox{oorange}{1.99} &
  \colorbox{rred}{\underline{1.61}} \\
\multicolumn{1}{l|}{MSE$\downarrow$} &
  \multicolumn{1}{c}{12.68} &
  \multicolumn{1}{c}{2.25} &
  \multicolumn{1}{c}{1.23} &
  \multicolumn{1}{c|}{1.34} &
  \multicolumn{1}{c}{-} &
  \multicolumn{1}{c}{4.58} &
  \multicolumn{1}{c}{{0.98}} &
  \multicolumn{1}{c}{\colorbox{oorange}{0.71}} &
  \multicolumn{1}{c}{\colorbox{rred}{\underline{0.50}}} \\
\multicolumn{1}{l|}{Grad$\downarrow$} &
  \multicolumn{1}{c}{{8.42}} &
  \multicolumn{1}{c}{15.10} &
  \multicolumn{1}{c}{12.97} &
  \multicolumn{1}{c|}{8.38} &
\multicolumn{1}{c}{-} &
  \multicolumn{1}{c}{29.78} &
  \multicolumn{1}{c}{\colorbox{oorange}{7.69}} &
  \multicolumn{1}{c}{{8.91}} &
  \multicolumn{1}{c}{\colorbox{rred}{\underline{7.14}}} \\
\multicolumn{1}{l|}{dtSSD$\downarrow$} &
  2.74 &
  2.28 &
  \multicolumn{1}{c}{2.04} &
  \multicolumn{1}{c|}{1.72} &
  - &
  2.41 &
  {1.77} &
  \colorbox{oorange}{1.65} &
  \colorbox{rred}{\underline{1.53}} \\ \hline
\end{tabular}
}
\end{center}
\vspace{-4mm}
\end{table*}
\subsection{Reference-Frame Strategy for Long Videos}
\label{subsec:ref_frame}
In long videos, the subject appearance can vary largely over time. Previous MatAnyone~\cite{yang2025matanyone}, which relies on a memory-propagation mechanism, often fails to matte out newly appearing human or clothing parts that were unseen in earlier frames.
This limitation arises because, during training, it is exposed only to short sequences (\eg, 8 frames), where only frames within local window are stored in memory (Fig.~\ref{fig:matanyone1vs2}(a)), preventing the model from modeling large appearance variations.
Simply extending the training window (\eg, 40 frames~\cite{lin2022rvm}) would drastically increase memory consumption.
To overcome this issue, we propose a reference-frame strategy that introduces long-range reference frames into memory \textit{beyond} the local training window (Fig.~\ref{fig:matanyone1vs2}(b)), following the training scheme of~\cite{zhou2023propainter}, thereby extending the temporal context without additional memory overhead.
To strengthen the reference-frame strategy, we further apply a random dropout augmentation that randomly masks patches in both RGB and alpha maps, which mitigates over-reliance on historical memory and enhances the model’s ability to handle unseen regions.
This strategy effectively simulates the challenging large appearance variations commonly encountered in long videos, leading to improved stability under challenging scenarios (see Fig.~\ref{fig:qualitative_long}).
\section{Experiments}
\label{sec:experiments}

\begin{figure*}[!t]
\begin{center}
    \vspace{-3mm}
    \includegraphics[width=\linewidth]{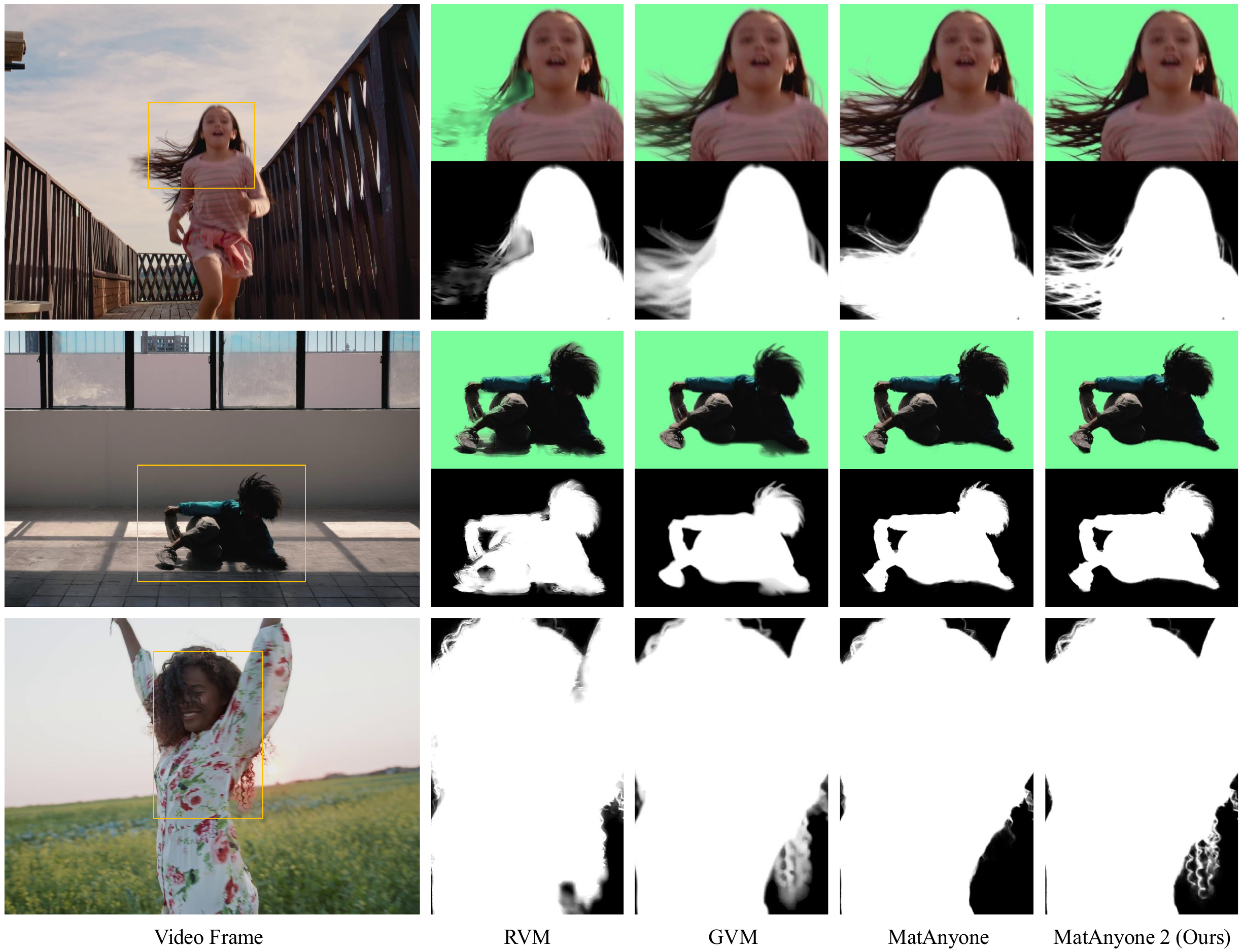}
    \vspace{-6mm}
    \caption{
    Qualitative comparisons on real-world videos.
    Our MatAnyone 2 significantly outperforms leading auxiliary-free (RVM~\cite{lin2022rvm}), diffusion-based (GVM~\cite{ge2025gvm}), and mask-guided (MatAnyone~\cite{yang2025matanyone}) approaches in both detail extraction and semantic accuracy under challenging conditions, \eg, wind-blown hair (first row) and complex lighting scenes (last two rows).
    }
    \vspace{-6mm}
\label{fig:qualitative_real}
\end{center}
\end{figure*}

\noindent \textbf{Implementation Details.}
Our MatAnyone~2 is trained on 8 A800-80G GPUs with a batch size of 16. Training clips are cropped to $480 \times 480$ and 8 frames. The matting backbone follows MatAnyone~\cite{yang2025matanyone}. With VMReal as the training set, no separate networks or losses for different data types are needed.
The Matting Quality Evaluator (MQE) is also trained on 8 A800-80G GPUs with a batch size of 16. Inspired by Depth Anything~\cite{yang2024depth}, the MQE uses pretrained DINOv3~\cite{simeoni2025dinov3} for feature extraction and a DPT~\cite{ranftl2021dpt} decoder for evaluation map prediction.

\subsection{Comparisons}
\label{subsuc:exp_vm}
We compare MatAnyone 2 with several state-of-the-art methods, including \textit{auxiliary-free (AF) methods}: MODNet~\cite{ke2022MODNet}, RVM~\cite{lin2022rvm}, RVM-Large~\cite{lin2022rvm}, and diffusion-based GVM~\cite{ge2025gvm}; as well as \textit{mask-guided methods}: AdaM~\cite{lin2023adam}, FTP-VM~\cite{huang2023ftp}, MaGGIe~\cite{huynh2024maggie}, and MatAnyone~\cite{yang2025matanyone}.

\subsubsection{Quantitative Evaluations}
\noindent\textbf{Synthetic Benchmarks.}
We evaluate on synthetic benchmarks using MAD and MSE for semantic accuracy, Grad~\cite{rhemann2009perceptually} for detail fidelity, Conn~\cite{rhemann2009perceptually} for perceptual quality, and dtSSD~\cite{erofeev2015perceptually} for temporal coherence. As shown in Table~\ref{tab:comparison_syn}, our method achieves the best scores across all four benchmarks, demonstrating strong semantic accuracy, detail quality, and temporal stability.
Compared with the leading mask-guided MatAnyone~\cite{yang2025matanyone}, our model reduces Grad and Conn by $27.1\%$ and $22.4\%$, validating the effectiveness of MQE-based video matting scaling. Moreover, despite GVM’s diffusion prior and MaGGIe’s per-frame instance masks, our CNN model with only first-frame masks still outperforms all methods, demonstrating the effectiveness of online matting-quality guidance and the VMReal dataset for video matting training.

\begin{table}[t]
\begin{center}
\setlength{\fboxsep}{3pt}
\caption{
    Quantitative comparisons on the CRGNN real-world dataset~\cite{wang2021crgnn}, where ground-truth alpha mattes are manually annotated every 10 frames on 19 videos. The best and second performances are marked in \colorbox{rred}{\underline{red}} and \colorbox{oorange}{orange}, respectively.
    }
\label{tab:crgnn_table}
\vspace{-3mm}
\renewcommand{\arraystretch}{1.1}
\renewcommand{\tabcolsep}{4.5mm}
\resizebox{\linewidth}{!} {
\begin{tabular}{l|cccc}
\hline
Methods          & MAD$\downarrow$ & MSE$\downarrow$ & Grad$\downarrow$ & dtSSD$\downarrow$ \\ \hline
\multicolumn{5}{l}{{\textit{Auxiliary-free}}}  \\ \hline
RVM~\cite{lin2022rvm}      & 5.98 & 2.79 & 13.68 & 5.36 \\
RVM-Large~\cite{lin2022rvm}     & 5.75 & 2.47 & \colorbox{oorange}{13.26} & 5.17 \\ 
GVM~\cite{ge2025gvm}      & \colorbox{oorange}{5.03} & \colorbox{oorange}{2.15} & 14.28 & \colorbox{oorange}{4.86} \\ \hline
\multicolumn{5}{l}{{\textit{Mask-guided}}}        \\ \hline
FTP-VM~\cite{huang2023ftp}     & 6.64 & 3.54 & 15.10 & 5.98 \\
MaGGIe~\cite{huynh2024maggie}      & 9.50 & 6.11 & 16.51 & 6.02 \\
MatAnyone & 5.76 & 3.04 & 15.55 & 5.44 \\
\textbf{Ours} & \colorbox{rred}{\underline{4.24}}  & \colorbox{rred}{\underline{2.00}}  & \colorbox{rred}{\underline{11.74}} & \colorbox{rred}{\underline{4.54}} \\ \hline
\end{tabular}
}
\end{center}
\vspace{-8mm}
\end{table}

\noindent\textbf{Real Benchmark.}
To assess model generalization on real-world videos, we further evaluate our method on the real-world CRGNN dataset~\cite{wang2021crgnn}, as shown in Table~\ref{tab:crgnn_table}. 
This dataset contains 19 real-world videos with ground-truth alpha mattes manually annotated every 10 frames. 
The substantial metric improvements over MatAnyone~\cite{yang2025matanyone} indicate enhanced semantic accuracy, finer details, and better temporal consistency.
Although GVM~\cite{ge2025gvm} achieves competitive results due to its strong diffusion prior, our MatAnyone~2 still delivers the best performance across all metrics, highlighting the superior generalizability and robustness of our scaling design under real-world conditions.

\subsubsection{Qualitative Evaluations}
Qualitative results on real-world videos are shown in Fig.~\ref{fig:qualitative_real}.
While the diffusion-based GVM~\cite{ge2025gvm} is robust across scenes, it produces blurry mattes with unnatural transitions, especially around hair. Mask-guided MatAnyone~\cite{yang2025matanyone} is also robust due to segmentation supervision, but its weak boundary constraints lead to segmentation-like edges.
In contrast, our method delivers superior semantic robustness and fine matting details under challenging real-world conditions, including wind-blown hair (\eg, first row) and complex lighting (\eg, last two rows), showing that MQE-guided supervision and VMReal offer strong, reliable signals for high-quality real-world video matting.

\subsubsection{MQE Performance Evaluations}
\label{subsuc:exp_qe}
For the MQE to provide effective guidance in both online training and offline selection, it should accurately identify erroneous regions in the predicted mattes.
As illustrated in Fig.~\ref{fig:qe_eval}, the MQE successfully highlights common failure cases such as (a) segmentation-like mattes along object boundaries and (b) semantic mispredictions between foreground and background.
These results demonstrate that the learned MQE provides a solid foundation for guiding video matting training and data curation. More evaluation and analysis are provided in the supplementary (see Sec.~\ref{suppl_subsec: qe_eval}).

\begin{table}[t]
\vspace{-2mm}
\caption{Ablation study of online guidance $\mathcal{L}_{eval}$, VMReal dataset, and reference-frame strategy on YouTubeMatte ($1920 \times 1080$). We employ MatAnyone~\cite{yang2025matanyone} as the baseline, \textit{i.e.}, Exp. (a).}
\centering
\vspace{-1mm}
\renewcommand{\arraystretch}{1.1}
\renewcommand{\tabcolsep}{2mm}
\resizebox{\linewidth}{!} {
\begin{tabular}{l|ccc|cccc}
\toprule
Exp. & $\mathcal{L}_{eval}$ & VMReal & Ref. Train & MAD$\downarrow$ & MSE$\downarrow$ & Grad$\downarrow$ & dtSSD$\downarrow$\\ 
\midrule
(a)   &   & & & 1.99 & 0.71 & 8.91 & 1.65 \\ \midrule
(b)  & \checkmark &   &    &  1.90  &  0.62 & 8.20 & 1.63 \\ 
(c) & \checkmark & \checkmark  &    & 1.76 &  0.61  & 7.65  & 1.54 \\ 
(d) &\checkmark & \checkmark  & \checkmark  &  {\bf 1.61} & {\bf 0.50} & {\bf 7.13} & {\bf 1.53} \\ \bottomrule
\end{tabular}
}
\label{tab:ablation}
\vspace{-2mm}
\end{table}

\subsection{Ablation Study}
\noindent \textbf{Effectiveness of Online Guidance $\mathcal{L}_{eval}$.}
To verify the effectiveness of MQE’s online guidance, we train our video matting model using the same schedule as MatAnyone, where matting and segmentation data are supervised by different losses (Fig.~\ref{fig:matanyone1vs2}(a)). Since the MQE can generate evaluation maps for both data types, we incorporate $\mathcal{L}_{eval}$ as an additional online quality-guidance loss.
As shown in Table~\ref{tab:ablation}, the improvement is most apparent in metrics related to semantic accuracy (MAD, MSE) and detail fidelity (Grad). This indicates that the MQE provides a more comprehensive and reliable assessment of both semantic correctness and fine details, offering stronger and more stable guidance than the previous weak boundary supervision.

\noindent \textbf{Enhancement from Training with VMReal.}
Beyond providing online guidance during training, we further employ MQE to construct a large-scale real-world video matting dataset, VMReal.
From Table~\ref{tab:ablation}(b) to (c), noticeable improvements are observed in MAD, Grad, and dtSSD, suggesting comprehensive gains not only in semantic robustness and boundary fidelity, but also in temporal consistency.
To further validate the benefit of VMReal, we also train RVM~\cite{lin2022rvm} on this dataset and observe consistent improvements across all metrics (Table~\ref{tab:rvm_vmreal} in the supplementary). 
These results demonstrate that VMReal serves as a strong and generalizable supervisory signal, boosting video matting performance across different models.

\begin{figure}[!t]
\begin{center}
    \vspace{-3mm}
    \includegraphics[width=\linewidth]{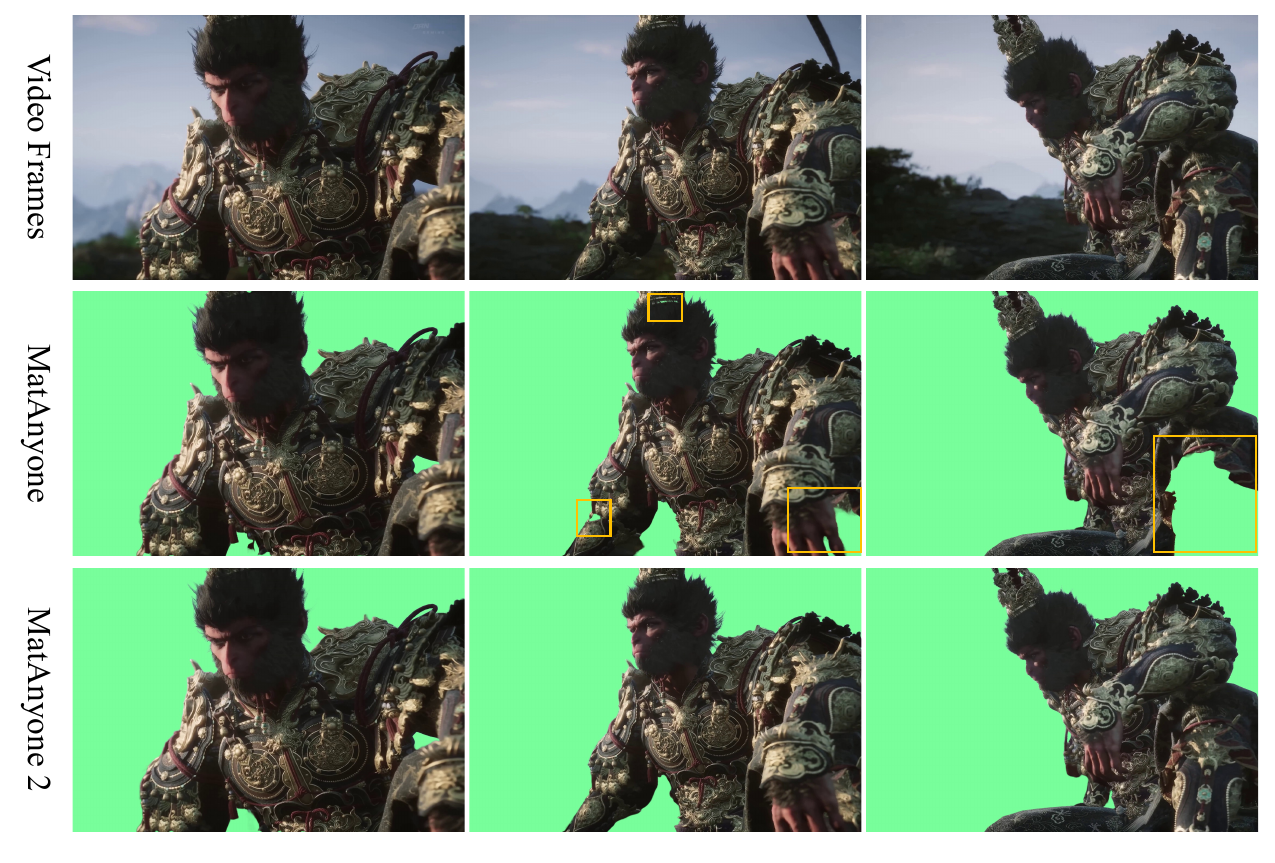}
    \vspace{-7mm}
    \caption{
    Effectiveness of the reference-frame strategy. MatAnyone~\cite{yang2025matanyone} struggles with newly appearing subject regions in long videos, whereas MatAnyone~2, trained with the reference-frame strategy, robustly identifies them.
    }
    \vspace{-7mm}
\label{fig:qualitative_long}
\end{center}
\end{figure}

\noindent \textbf{Effectiveness of Reference-Frame Strategy for Large Variations.}
With the training sequence length kept unchanged, we adopt the reference-frame training strategy with random dropout augmentation in the data pipeline. As shown in Table~\ref{tab:ablation}(d), this yields further improvements, especially in semantic accuracy (MAD and MSE), indicating enhanced semantic robustness without incurring additional memory cost.
This improved capability allows the model to better handle large appearance changes across long videos, such as the emergence of new body or clothing parts that do not appear in earlier frames (see Fig.~\ref{fig:qualitative_long}).

\section{Conclusion}
\label{sec:conclusion}
\vspace{-1mm}
We presented MatAnyone 2, a next-generation video matting framework that integrates a learned Matting Quality Evaluator (MQE) for scalable training.
The MQE jointly assesses semantic accuracy and boundary detail without ground truth, serving as both an online guidance signal and an offline data selection module.
Leveraging MQE, we further constructed VMReal, the first large-scale, real-world video matting dataset with diverse scenes and high-quality annotations.
In addition, a reference-frame training strategy was introduced to handle large appearance variations in long videos.
Extensive experiments show that MatAnyone 2 achieves leading performance on both synthetic and real-world benchmarks, significantly improving semantic stability, boundary fidelity, and generalization in real scenarios.

\clearpage

\vspace{2mm}
\noindent{\bf Acknowledgement.} This research is supported by cash and in-kind funding from NTU S-Lab and industry partner(s).

{
    \small
    \bibliographystyle{ieeenat_fullname}
    \bibliography{main}
}

\clearpage
\onecolumn

\begin{center}
	\Large\textbf{{Appendix}}\\
	\vspace{8mm}
\end{center}
In this supplementary material, we provide additional discussions and results to supplement the main paper. 
In Sec.~\ref{sec:training_details}, we present the training details of our MatAnyone 2, which includes the Matting Quality Evaluator (MQE) model and the video matting model.
In Sec.~\ref{sec:vmreal}, we discuss more details about \textit{VMReal}, the large-scale, real-world video matting dataset we constructed, including an overview, the data curation pipeline, and some examples for demonstration.
We include comprehensive results in Sec.~\ref{sec:more_results} to further show our performance, including those for annotation branches, ablation study, MQE performance, and qualitative comparisons.
We also include a demo video in \href{https://pq-yang.github.io/projects/MatAnyone2/}{[project page]} to showcase more results on real-world cases in videos.
Finally, we include a brief discussion on the limitations of our method in Sec.~\ref{sec:limitations}.


\renewcommand\thesection{\Alph{section}}
\setcounter{section}{0}
{   
    \hypersetup{linkcolor=blue}
    \tableofcontents
}

\clearpage
\section{More Training Details}
\label{sec:training_details}
\subsection{Matting Quality Evaluator Model}
\noindent \textbf{Training Data.}
As mentioned in Sec.~\ref{subsec:qe_model} in the manuscript, we formulate MQE learning as a binary segmentation task that predicts whether each pixel of an alpha matte is erroneous (class 0) or reliable (class 1). Since no ground-truth evaluation maps naturally exist for this purpose, we construct pseudo ground-truth labels using the P3M-10k~\cite{li2021p3m} dataset, which provides high-quality human-annotated alpha mattes corresponding to real-world images.
To generate $M^{eval}_{gt}$, we evaluate a predicted alpha $\hat{\alpha}$ against the ground-truth $\alpha_{gt}$ by computing a patch-level discrepancy score. Specifically, each image is divided into a $7 \times 7$ grid of non-overlapping patches, and for each patch we compute a weighted combination of standard matting metrics:
\begin{equation}
    \mathcal{D}(\cdot) = 0.9 \cdot MAD + 0.1 \cdot Grad.
\end{equation}
All discrepancy values are then min--max normalized into the range $[0,1]$. The binary evaluation map is defined as
\begin{equation}
    M^{eval}_{gt} = \mathbb{I}\!\left( \mathcal{D}(\alpha_{gt}, \hat{\alpha}) < \delta \right),
\end{equation}
where $\mathbb{I}(\cdot)$ denotes the indicator function and we empirically set the threshold $\delta=0.2$. This process effectively captures both semantic- and detail-level deviations, and naturally emphasizes boundary regions where most matting errors occur.

\noindent \textbf{Loss Functions.}
Since reliable pixels dominate erroneous ones, the training data exhibits a strong class imbalance. To address this, we adopt focal loss~\cite{lin2017focal} to emphasize hard erroneous regions and stabilize the training of the MQE model:
\begin{equation}
\mathcal{L}_{\text{focal}} =
- \alpha (1 - p)^{\gamma} y \log(p)
- (1 - \alpha) p^{\gamma} (1 - y)\log(1 - p),
\end{equation}
where $p$ denotes the predicted probability for class~1 (reliable pixel) and 
$y \in \{0,1\}$ is the corresponding ground-truth label from $M^{eval}_{gt}$.
We set $\gamma = 2$ and $\alpha = 0.25$ following~\cite{lin2017focal}.
In addition, we also employ the dice loss $\mathcal{L}_{dice}$ following~\cite{yang2021dice}.
The overall MQE training loss is summarized as:
\begin{equation}
\mathcal{L}_{MQE}
= \mathcal{L}_{\text{focal}}
+ \mathcal{L}_{\text{dice}}.
\end{equation}

\noindent \textbf{Implementation Details.}
MQE is trained on 8 A800-80G GPUs with a batch size of 16. Inspired by Depth Anything~\cite{yang2024depth}, the MQE uses pretrained DINOv3~\cite{simeoni2025dinov3} for feature extraction and a DPT~\cite{ranftl2021dpt} decoder for evaluation map prediction.
We train MQE for 80K iterations, with a learning rate of $3 \times 10^{-6}$ for encoder and $10\times$ larger for decoder.

\subsection{Video Matting Model}
\noindent \textbf{Augmentations for Reference Frames.}
As discussed in Sec.~\ref{subsec:ref_frame} of the manuscript, we introduce a reference-frame strategy that samples frames outside the local training window into the memory bank, enabling the model to observe larger appearance variations without increasing memory overhead. However, this sampling is probabilistic and does not guarantee that the selected reference frames exhibit sufficiently large appearance differences from the current window. 
To mitigate this issue, we introduce a random dropout augmentation on reference frames. Since unseen regions with respect to previous frames typically appear near object boundaries (see Fig.~\ref{fig:qualitative_long} in the manuscript), we randomly sample $0\sim3$ patches from boundary regions and $0\sim1$ patches from non-boundary regions on reference frames. Each patch is randomly sized within the range of $[50, 100]$. For both RGB and alpha matte of the reference frame, the corresponding patch areas are masked out by setting their values to zero. 
This augmentation further enables the reference-frame strategy to better simulate large appearance variations in long videos, thereby enhancing its effectiveness in handling unseen or highly varying regions.

\noindent \textbf{Loss Functions.}
Following ~\cite{lin2022rvm}, matting losses $\mathcal{L}_{mat}$ commonly include: (1) L1 loss for semantics $\mathcal{L}_{l1}$, (2) pyramid Laplacian loss~\cite{hou2019context} for matting details $\mathcal{L}_{lap}$, and (3) temporal coherence loss~\cite{sun2021deep} $\mathcal{L}_{tc}$ for flickering reduction.
When training with our VMReal dataset consisting of triplets of $\langle I_{rgb}, \alpha, M^{eval} \rangle$, where $M^{eval} \in \{0,1\}^{H \times W}$ is an evaluation map with class~1 indicates reliable pixels,
we convert the above losses into a masked version $\mathcal{L}_{mat}^M$ that only supervises reliable regions.
For frame $t$, suppose we have the predicted alpha matte $\hat{\alpha}$ w.r.t. its ground-truth (GT) $\alpha$.
Let $R_t = \mathbb{I}(M^{eval}_t = 1)$ denote the reliability mask and $\odot$ the element-wise product.
The masked L1 loss and Laplacian loss are defined as:
\begin{equation}
    \mathcal{L}_{l1}^M 
    =
    \frac{\big\| R_t \odot (\hat{\alpha}_t - \alpha_t) \big\|_1}
    {\| R_t \|_1 + \epsilon},
\end{equation}
\begin{equation}
    \mathcal{L}_{lap}^M
    =
    \sum_{s=1}^{5} \frac{2^{s-1}}{5} \,
    \frac{\big\| R_t \odot \big( L_{pyr}^s(\hat{\alpha}_t) - L_{pyr}^s(\alpha_t) \big) \big\|_1}
    {\| R_t \|_1 + \epsilon}.
\end{equation}
For temporal coherence, we only consider pixels that are reliable in both consecutive frames.
Let $R_t^{tc} = R_t \odot R_{t-1}$ and define  
$\Delta\hat{\alpha}_t = \hat{\alpha}_t - \hat{\alpha}_{t-1}$ and  
$\Delta\alpha_t = \alpha_t - \alpha_{t-1}$.  
Then the masked temporal coherence loss is:
\begin{equation}
\mathcal{L}_{tc}^M =
\frac{
\big\| R_t^{tc} \odot (\Delta\hat{\alpha}_t - \Delta\alpha_t) \big\|_2^2
}{
\| R_t^{tc} \|_1 + \epsilon
}.
\end{equation}
In summary, the masked matting loss is:
\begin{equation}
    \mathcal{L}_{mat}^M
    = \mathcal{L}_{l1}^M + \mathcal{L}_{lap}^M + \mathcal{L}_{tc}^M.
\end{equation}
In addition to the standard matting losses, in Sec.~\ref{subsec:offline} of the manuscript, we further introduce an online matting-quality guidance enabled by the MQE model. The MQE provides a pixel-wise error probability map $P^{(0)}_{eval}$ for each predicted alpha, offering on-the-fly quality feedback during training (Fig.~\ref{fig:matanyone1vs2}(b)). This map serves as a penalty mask that suppresses unreliable regions, and the corresponding guidance loss is defined as
\begin{equation}
    \mathcal{L}_{eval} = \| P^{(0)}_{eval} \|_1,
\end{equation}
which encourages lower predicted error probabilities and thus more accurate alpha predictions.
Finally, the overall video matting training loss is
\begin{equation}
    \mathcal{L}_{mat}^{total}
    = \mathcal{L}_{mat}^{M} + 0.1\,\mathcal{L}_{eval}.
\end{equation}
Empirically, we set the weight of $\mathcal{L}_{eval}$ to 0.1 so that its magnitude remains comparable to the main matting loss $\mathcal{L}_{mat}^{M}$, leading to stable optimization during training.

\section{VMReal Dataset}
\label{sec:vmreal}
\subsection{Overview}
\begin{figure}[h]
\begin{center}
    \vspace{-0mm}
    \includegraphics[width=\linewidth]{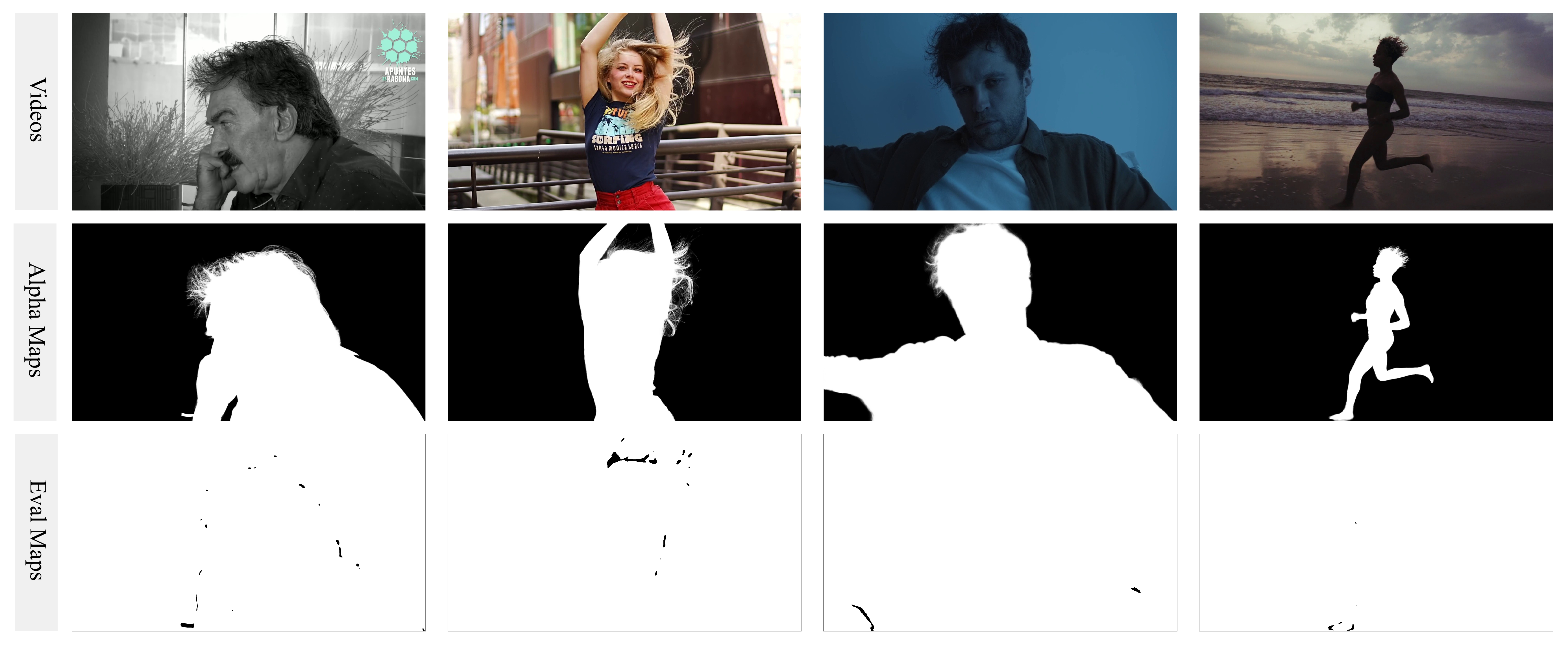}
    \vspace{-2mm}
    \caption{
    Examples from our VMReal dataset.
    VMReal contains diverse and challenging real-world video matting cases, including close-up portraits, large-motion hair dynamics, and low-light environments. For each video, we provide high-quality alpha mattes together with the corresponding evaluator maps, enabling reliable supervision and assessment under complex real-world conditions.
    }
    \vspace{-8mm}
\label{fig:gallery}
\end{center}
\end{figure}

As summarized in Table~\ref{tab:vm_datasets}, our \textit{VMReal} dataset is substantially larger and more realistic than existing video matting datasets. 
Whereas prior datasets such as VM108~\cite{zhang2021vm108}, VideoMatte240K~\cite{lin2021bgm}, VM800~\cite{yang2025matanyone}, and SynHairMan~\cite{ge2025gvm} are relatively small in scale and are purely synthetic composites of foregrounds and backgrounds, \textit{VMReal} comprises 28K real-world video clips with a total of 2.4M frames. 
To the best of our knowledge, this makes \textit{VMReal} the first large-scale real-world video matting dataset that provides diverse scenes together with high-quality annotations.
The dataset spans a wide range of human-centric variations, including:  
\emph{(i) subject count}: from single-person to multi-person scenarios;  
\emph{(ii) appearance diversity}: individuals of different genders, ages, and nationalities;  
\emph{(iii) lighting conditions}: natural light, indoor lighting, backlight, and other challenging illumination setups;  
\emph{(iv) motion patterns}: both human–human and human–object interactions.  
Among them, many are challenging cases for the video matting task, as shown in Fig.~\ref{fig:gallery}.

\begin{table}[h]
\caption{
\textbf{Comparison on Datasets.}
We compare existing video matting (VM) datasets for training, in terms of the number of video clips, the number of total frames, and whether the input frames are synthesized (foreground-background composition) or from real videos.
Whereas prior datasets such as VM108~\cite{zhang2021vm108}, VideoMatte240K~\cite{lin2021bgm}, VM800~\cite{yang2025matanyone}, and SynHairMan~\cite{ge2025gvm} are relatively small in scale and are purely synthetic composites of foregrounds and backgrounds, \textit{VMReal} comprises 28K \textit{real-world} video clips with a total of 2.4M frames. 
}
\centering
\vspace{-0mm}
\renewcommand{\arraystretch}{1.2}
\renewcommand{\tabcolsep}{2.0mm}
\scalebox{0.85}{
    \begin{tabular}{l|ccccc}
    \toprule
    Datesets & VM108~\cite{zhang2021vm108} & VideoMatte240K~\cite{lin2021bgm} & VM800~\cite{yang2025matanyone} & SynHairMan~\cite{ge2025gvm} & \cellcolor{cvprblue!10}\textbf{VMReal (ours)} \\ 
    \midrule
    \#clips & 108 & 484 & 826 & 200 & \cellcolor{cvprblue!10}28K   \\ 
    \#frames & 92K & 240K & 320K & 18K & \cellcolor{cvprblue!10}2.4M \\ 
    Realism & \text{\sffamily x} & \text{\sffamily x} & \text{\sffamily x} & \text{\sffamily x} & \cellcolor{cvprblue!10}\checkmark \\
    \bottomrule
    \end{tabular}
}
\label{tab:vm_datasets}
\vspace{-3mm}
\end{table}

\subsection{Data Curation Pipeline}
\noindent \textbf{Data Collection.}
Our data are primarily collected from two sources. 
(1) A high-quality subset is collected from online video-footage platforms and YouTube, consisting of 4.5K 1080p videos with rich appearance details. 
(2) The remaining videos are derived from the large-scale SA-V~\cite{ravi2024sam2} dataset, where we filter out a human-centric subset, typically at 720p with diverse scenes, motions, and lighting conditions. 

\noindent \textbf{Data Preprocessing.}
Since SA-V~\cite{ravi2024sam2} does not provide semantic labels for each mask and contains many partial or fragmented masks (\eg, only head or only upper body), we apply a two-stage filtering-and-merging pipeline to obtain complete human masks.
\textit{(i) Reducing redundant masks.}
We first focus on salient objects by merging masks provided by manual annotations with those from automatic annotations. 
Redundant masks are removed through an iterative merging procedure. 
Specifically, a mask $M_i$ is considered redundant if it is almost completely covered by another mask $M_j$, i.e.,
\[
IoU(M_i, M_j) = \frac{|M_i \cap M_j|}{|M_i|} > 0.9.
\]
\textit{(ii) Human-mask identification and grouping.}
After removing redundant masks, we perform human-mask nomination using OWL-SAM~\cite{huggingface:owlsam} with the text prompt \texttt{"person"}. 
Each remaining mask is then grouped to the human instance it belongs to, and grouped masks are merged into a single complete human mask. 
Masks that cannot be assigned to any person are treated as non-human and discarded. 
Following this merging step, fragmented or incomplete masks are also removed, aiming to retain complete human masks suitable for matting.

\begin{figure}[h]
\begin{center}
    \vspace{-0mm}
    \includegraphics[width=\linewidth]{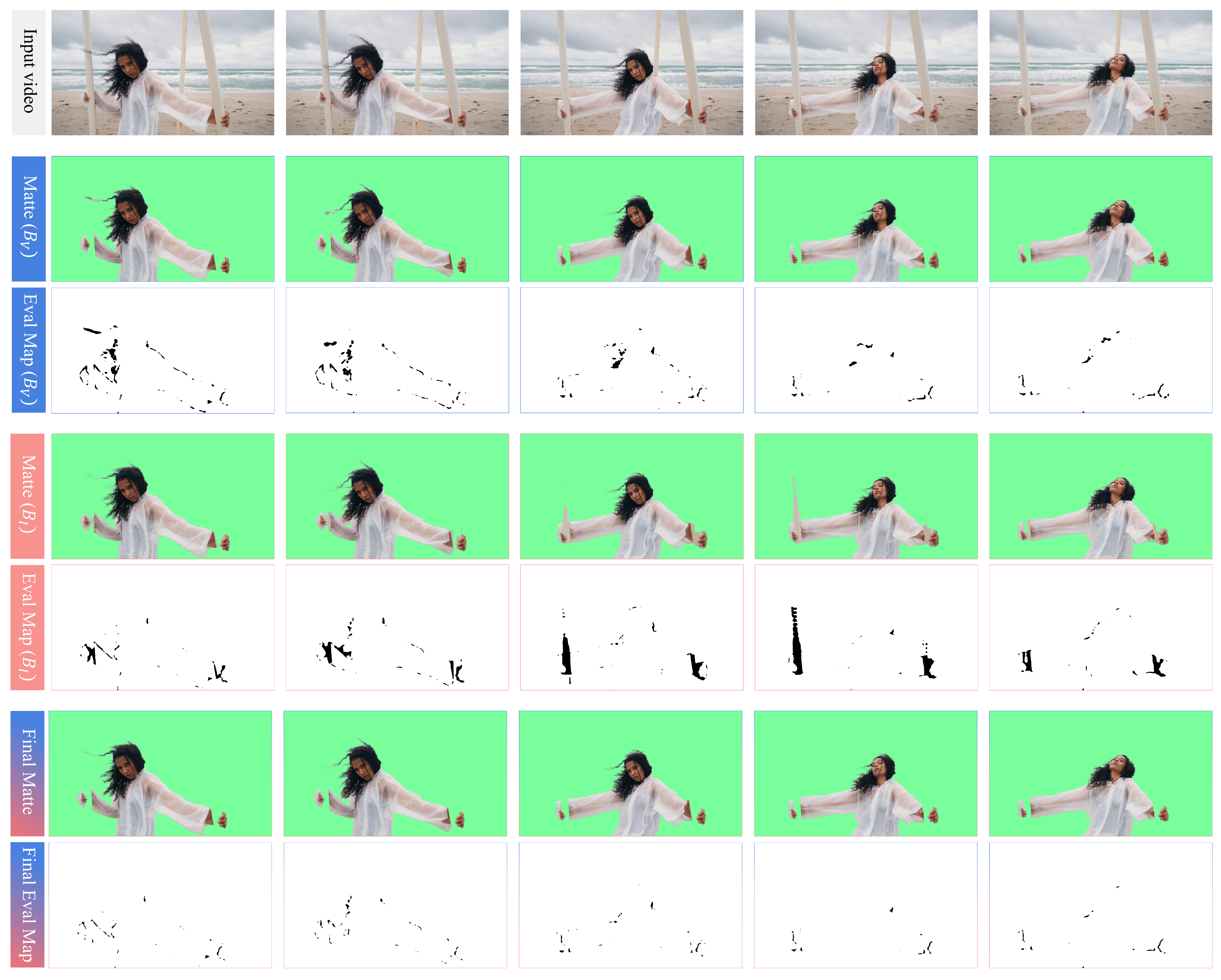}
    \vspace{-4mm}
    \caption{
    Illustration of our dual-branch annotation pipeline on a video sequence.
    For the video-based branch $B_V$, most errors concentrate on fine-grained hair regions where the model produces segmentation-like mattes, while the overall semantic structure remains stable. In contrast, the image-based branch $B_I$ exhibits highly unstable semantics across frames (e.g., the two poles held in both hands), and although it often recovers sharp details, these details may fluctuate noticeably even when the subject barely moves (e.g., degraded hair details in the second frame).
    Motivated by these complementary behaviors, we adopt a dual-branch annotation design: we use the stable and temporally consistent predictions from $B_V$ as base, while selectively integrating high-quality boundary details from $B_I$ guided by MQE. The merged final matte achieves a better balance between semantics and fine details, and the corresponding evaluation maps show a significant reduction in error pixels, indicating a decent quality of the final matte.
     \textbf{(Zoom in for best view)}
    }
    \vspace{-6mm}
\label{fig:im_branch}
\end{center}
\end{figure}

\noindent \textbf{Automated Dual-branch Annotation.}
As illustrated in Fig.~\ref{fig:data_pipeline}, two complementary matting branches are used for annotations: the temporally stable video matting branch $B_V$ and the detail-preserving image matting branch $B_I$.
\textit{(i) Automatic Labeling.}
Given an input video, we first obtain per-frame segmentation masks using SAM~2~\cite{ravi2024sam2}. 
These masks serve different purposes for the two annotation branches. 
For the video matting branch $B_V$, only the first-frame segmentation mask is required: it specifies the target subject, and the model refines and subsequently propagates the target-aware matte to all frames, producing a temporally stable alpha sequence $\alpha_V$. 
In contrast, the image matting branch $B_I$ requires target localization for every frame. 
We convert each frame's SAM~2 mask into bounding boxes and points, which are then used as target prompts for $B_I$ to generate high-quality alpha mattes $\alpha_I$ with fine boundary details. 
\textit{(ii) Automatic Evaluation.}
Both $\alpha_V$ and $\alpha_I$ are evaluated by the trained MQE, which takes the per-frame SAM~2 segmentation mask as semantic cues. 
The MQE produces pixel-wise error probability maps $M_V^{eval}$ and $M_I^{eval}$ without requiring ground-truth alpha mattes. 
In practice, errors in $M_V^{eval}$ are mostly concentrated in fine structures such as hair strands, reflecting the weaker boundary fidelity of $B_V$. 
In contrast, $M_I^{eval}$ exhibits semantic inconsistency errors over both boundary and non-boundary regions due to the temporal instability of $B_I$, and its error distribution varies across consecutive frames, as shown in Fig.~\ref{fig:im_branch} (pink rows).
\textit{(iii) Soft Blending.}
We take the stable and temporally consistent predictions from $B_V$ as the base, while MQE selectively integrates high-quality boundary details from $B_I$ where $B_V$ is deemed erroneous ($M^{eval}_{V}=0$) but $B_I$ is reliable ($M^{eval}_{I}=1$). We denote such regions as fusion mask $M^{F}=M^{eval}_{I} \odot (1 - M^{eval}_{V})$.
To avoid boundary artifacts, we smooth the fusion mask $M^{F}$ with a Gaussian blur (kernel size $9 \times 9$, $\sigma = 5.0$), resulting in seamless and soft blending between the two branches.
The final refined alpha matte is obtained by:
\begin{equation}
\alpha = \alpha_{V} \odot (1-M^{F}) + \alpha_{I} \odot M^{F},
\end{equation}
which preserves the semantic stability of $B_V$ while incorporating fine boundary details from $B_I$.
Meanwhile, the fused evaluation map is updated as $M^{eval}=M^{eval}_{V}\cup M^{eval}_{I}$.
Note that for the videos from the SA-V dataset, our two branches are video segmentation (SA-V annotations) and video matting. Since SA-V videos contain limited boundary details, introducing an image-matting branch would not provide additional benefits.

\begin{table}[h]
\begin{center}
\caption{
    Quantitative comparisons on the image matting branch $B_I$ (image matting model MattePro~\cite{github:mattepro} guided by per-frame segmentation masks from SAM 2~\cite{ravi2024sam2}) and video matting branch $B_V$ (MatAnyone~\cite{yang2025matanyone}).
    For both low and high resolutions, the performance of $B_I$ is significantly inferior to that of the video matting branch $B_V$, reflecting $B_I$'s susceptibility to per-frame fluctuations and semantic inconsistency.
}
\label{tab:im_branch}
\vspace{-0mm}
\renewcommand{\arraystretch}{1.3}
\renewcommand{\tabcolsep}{4.0mm}
\resizebox{0.98\linewidth}{!}{
\begin{tabular}{l|ccccc|cccc}
\hline
\multirow{2}{*}{Method} &
\multicolumn{5}{c|}{{VideoMatte (512 × 288)}} &
\multicolumn{4}{c}{{VideoMatte (1920 × 1080)}} \\ \cline{2-10} 
 & MAD$\downarrow$ & MSE$\downarrow$ & Grad$\downarrow$ & dtSSD$\downarrow$ & Conn$\downarrow$ & MAD$\downarrow$ & MSE$\downarrow$ & Grad$\downarrow$ & dtSSD$\downarrow$ \\ \hline \hline
 $B_I$\scriptsize(SAM~2~\cite{ravi2024sam2}+MattePro~\cite{github:mattepro}) &
  8.96 &
  2.38 &
  2.02 &
  2.36 &
  0.68 &
  5.01 &
  0.67 &
  5.97 &
  2.24 \\
 $B_V$\scriptsize(MatAnyone~\cite{yang2025matanyone}) &
  \textbf{5.15} &
  \textbf{0.93} &
  \textbf{0.67} &
  \textbf{1.18} &
  \textbf{0.26} &
  \textbf{4.24} &
  \textbf{0.33} &
  \textbf{4.00} &
  \textbf{1.19} \\
  \hline
\end{tabular}
}
\end{center}
\end{table}
\section{More Results}
\label{sec:more_results}
\subsection{Performance of Image Matting Branch on Video Data}
Although the image matting branch $B_I$ produces sharper and more detailed boundaries, it struggles to maintain temporal stability and semantic consistency when applied to video data, as illustrated in Fig.~\ref{fig:im_branch}. 
To further quantify this behavior, we evaluate $B_I$ (image matting model~\cite{github:mattepro} guided by per-frame segmentation masks from SAM 2~\cite{ravi2024sam2}) on standard video matting benchmarks. 
As shown in Table~\ref{tab:im_branch}, for both low and high resolutions, the performance of $B_I$ is significantly inferior to that of the video matting branch $B_V$ (video matting model MatAnyone~\cite{yang2025matanyone}), reflecting $B_I$'s susceptibility to per-frame fluctuations and semantic drift. 
These results confirm that while $B_I$ provides high-fidelity details, it alone is not sufficient for stable video matting, motivating our dual-branch annotation design where the temporally consistent and semantically stable predictions from $B_V$ serve as the base, and high-quality boundary details from $B_I$ are selectively integrated under the guidance of the MQE.

\subsection{Improvement of Training RVM with VMReal}
As discussed in Sec.~\ref{subsec:offline} of the manuscript and Section~\ref{sec:vmreal} in this supplementary, our VMReal dataset provides substantially larger scale and higher realism compared with existing video matting datasets. 
Beyond the quantitative gains of our model from VMReal in Table~\ref{tab:ablation} of the manuscript, we further evaluate the effectiveness of VMReal by training another video matting model, RVM~\cite{lin2022rvm} on it.
From Table~\ref{tab:rvm_vmreal}, it can be observed that consistent improvements show across all metrics on both low- and high-resolution benchmarks. This confirms that VMReal provides a strong and generalizable supervisory signal that benefits different video matting models. 

\begin{table}[h]
\begin{center}
\caption{
    Quantitative comparisons on the RVM~\cite{lin2022rvm} model when trained with the VMReal dataset. 
    The comparison clearly shows that training with VMReal yields consistent improvements across all metrics at both resolutions.
}
\label{tab:rvm_vmreal}
\vspace{0mm}
\renewcommand{\arraystretch}{1.2}
\renewcommand{\tabcolsep}{3.0mm}
\scalebox{0.85}{
\begin{tabular}{l|ccccc|cccc}
\hline
\multirow{2}{*}{Method} &
\multicolumn{5}{c|}{{YouTubeMatte (512 × 288)}} &
\multicolumn{4}{c}{{YouTubeMatte (1920 × 1080)}} \\ \cline{2-10} 
 & MAD$\downarrow$ & MSE$\downarrow$ & Grad$\downarrow$ & dtSSD$\downarrow$ & Conn$\downarrow$ & MAD$\downarrow$ & MSE$\downarrow$ & Grad$\downarrow$ & dtSSD$\downarrow$ \\ \hline \hline
RVM\scriptsize(\textbf{w/o VMReal}) &
  4.08 &
  1.97 &
  1.34 &
  1.81 &
  0.60 &
  4.27 &
  2.25 &
  15.10 &
  2.28 \\
RVM\scriptsize(\textbf{w/ VMReal}) &
  {3.32}\scriptsize(\textcolor{red}{-0.76}) &
  1.34\scriptsize(\textcolor{red}{-0.63}) &
  1.20\scriptsize(\textcolor{red}{-0.14}) &
  1.79\scriptsize(\textcolor{red}{-0.02}) &
  0.49\scriptsize(\textcolor{red}{-0.11}) &
  3.37\scriptsize(\textcolor{red}{-0.9}) &
  1.51\scriptsize(\textcolor{red}{-0.74}) &
  14.40\scriptsize(\textcolor{red}{-0.70}) &
  2.22\scriptsize(\textcolor{red}{-0.06}) \\
  \hline
{MatAnyone 2} &
  {{2.30}} &
  {{0.78}} &
  {{0.78}} &
  {{1.45}} &
  {{0.32}} &
  {{1.61}} &
  {{0.50}} &
  {{7.13}} &
  {{1.53}} \\
  \hline
\end{tabular}
}
\end{center}
\vspace{-4mm}
\end{table}

\subsection{Matting Quality Evaluator Performance}
\label{suppl_subsec: qe_eval}
\begin{figure}[h]
\begin{center}
    \includegraphics[width=0.5\linewidth]{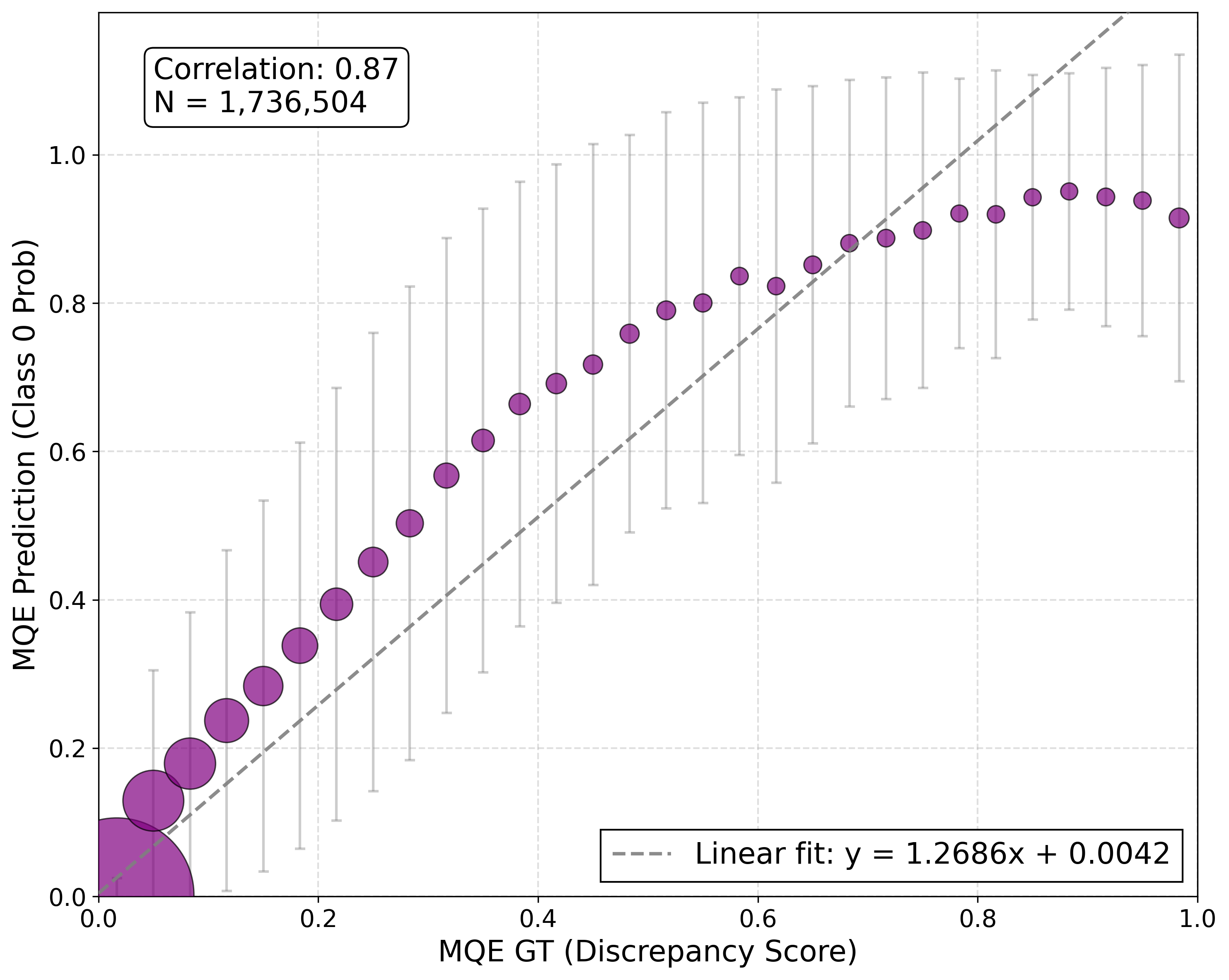}
    \vspace{-1mm}
    \caption{
    Correlation between MQE ground truth and prediction.
    We visualize the mean MQE prediction over 30 binned GT discrepancy intervals, with bubble size indicating the number of patches in each bin and error bars showing the standard deviation. The MQE prediction increases monotonically with the GT discrepancy and saturates for highly unreliable regions (Pearson $r = 0.87$). This saturation occurs because the evaluator becomes more conservative when encountering highly unreliable patches, pushing more pixels toward the “error’’ class to avoid missing true errors. Since erroneous pixels constitute only a small fraction of the data, this behavior preserves most valid supervision while ensuring that high-discrepancy regions are consistently flagged as unreliable.
        }
    \vspace{-5mm}
\label{fig:qe_corr}
\end{center}
\end{figure}

\noindent \textbf{Correlation Analysis.} 
To further examine the performance of our Matting Quality Evaluator (MQE), we visualize the relationship between the MQE ground-truth (GT) discrepancy score and the model’s predicted class~0 probability. 
As mentioned in Sec.~\ref{subsec:qe_model} in the manuscript, we compute patch-level GT discrepancy scores $\mathcal{D}(\cdot)=0.9 \cdot MAD+0.1 \cdot Grad$ considering both semantic-level and detail-level deviations.
We also obtain MQE predictions by averaging the per-pixel class~0 probability within each patch. 
All patch pairs are then grouped into 30 uniformly spaced bins along the GT discrepancy axis. For each bin, we plot the mean MQE prediction together with its standard deviation, while the bubble size indicates the number of patches falling into that bin. This aggregated visualization provides a clear and robust view of the MQE’s behavior over a large number of samples (1,736,504 in total).
The resulting curve (Fig.~\ref{fig:qe_corr}) demonstrates that our MQE predictions exhibit a \textit{strong and positive correlation} with the GT discrepancy score (Pearson $r = 0.87$), validating its ability to reliably reflect patch-level error extent. 
In the low-to-mid discrepancy range ($\mathcal{D}(\cdot) < 0.5$), the MQE response grows in an approximately linear manner, indicating that the evaluator is well calibrated and sensitive to subtle variations in patch quality. When the discrepancy becomes large ($\mathcal{D}(\cdot) > 0.5$), the prediction gradually saturates toward 1.0, showing a conservative tendency to flag more pixels as erroneous in highly unreliable regions. 
Since erroneous pixels constitute only a small portion of the training data, this saturation behavior does not affect the amount of valid supervisory signal much. 
Instead, it helps ensure that severe errors are not overlooked, aligning with the design motivation of using MQE as a reliable estimator for guiding large-scale video matting training.

\noindent \textbf{MQE-based Metrics.}
For real-world videos without ground-truth alpha mattes, reference-based metrics can only be computed within the core regions~\cite{yang2025matanyone} (left part of Table~\ref{tab:semantic_eval_metric}).
These metrics (MAD, MSE, dtSSD) mainly assess semantic accuracy in core regions, where segmentation masks predicted by SAM~2~\cite{ravi2024sam2} are treated as pseudo ground truth.
We observe that our method matches the semantic accuracy of MatAnyone~\cite{yang2025matanyone} while achieving better temporal stability, demonstrating that it effectively learns robust segmentation priors from the VMReal dataset.
Unlike MatAnyone, which often trades matting fidelity for semantic consistency, our qualitative results (Fig.~\ref{fig:teaser} and Fig.~\ref{fig:qualitative_real}) show that our method retains fine details while maintaining stable semantics.
However, without ground truths, reference-based metrics fail to assess boundary regions, which are crucial for matting quality.
To address this, we leverage our learned MQE to produce a pixel-wise evaluation map that predicts whether each pixel is erroneous.
From this map, we define three metrics (right part of Table~\ref{tab:semantic_eval_metric}):
ERR, the overall proportion of pixels flagged as unreliable;
MER, the unreliable proportion restricted to the foreground region;
and BER, the unreliable proportion in a boundary band obtained via morphological dilation and erosion.
Because these metrics rely only on the MQE’s evaluation map, they provide \textit{non-reference} indicators of real-world matting quality when alpha annotations are unavailable. Their relative trends also generally align with those of the reference-based metrics reported in Table~\ref{tab:comparison_syn}.

\begin{table*}[h]
\vspace{0mm}
\begin{center}
\setlength{\fboxsep}{3pt}
\caption{
    Quantitative comparisons on a real-world benchmark without ground-truth alpha annotations~\cite{yang2025matanyone}. 
    ERR reflects the overall unreliable proportion across the entire frame;  
    MER evaluates the unreliable proportion restricted to the foreground region;  
    and BER focuses on the boundary band obtained through morphological dilation and erosion. 
    These evaluator-based quality metrics complement Table~\textcolor{cvprblue}{2} in the main manuscript, where reference-based metrics can only be computed within the core regions. 
    The best and second-best performances are highlighted in \colorbox{rred}{\underline{red}} and \colorbox{oorange}{orange}, respectively. 
    All evaluator-based metrics are reported as percentages.
}
\label{tab:semantic_eval_metric}
\vspace{-2mm}
\renewcommand{\arraystretch}{1.3}
\renewcommand{\tabcolsep}{8.0mm}
\scalebox{0.84}{
\begin{tabular}{l|ccc|ccc}
\hline
\multirow{2}{*}{Methods} & \multicolumn{3}{c|}{\textbf{Semantic Accuracy}} & \multicolumn{3}{c}{\textbf{Evaluator-based Quality}} \\ \cline{2-7}
 & MAD$\downarrow$ & MSE$\downarrow$ & dtSSD$\downarrow$ & ERR$\downarrow$ & MER$\downarrow$ & BER$\downarrow$ \\ 
\hline
\multicolumn{7}{l}{{\textit{Auxiliary-free}}} \\ \hline
MODNet~\cite{ke2022MODNet} & 16.47 & 14.68 & 3.16 & 2.31 & 2.20 & 27.11 \\
RVM~\cite{lin2022rvm} & 1.71 & 1.30 & 1.39 & 1.05 & 2.34 & 25.45 \\
RVM-Large~\cite{lin2022rvm} & \colorbox{oorange}{1.02} & \colorbox{oorange}{0.57} & 1.28 & 1.04 & 2.48 & 26.61 \\
GVM~\cite{ge2025gvm} & 10.65 & 10.35 & 1.26 & 1.31 & 1.81 & 24.53 \\ 
\hline
\multicolumn{7}{l}{{\textit{Masked-guided}}} \\ \hline
FTP-VM~\cite{huang2023ftp} & 2.89 & 2.29 & 1.63 & 1.43 & 4.06 & 30.76 \\
MaGGIe~\cite{huynh2024maggie} & 1.89 & 1.48 & 1.63 & 0.98 & 2.23 & 22.06 \\
MatAnyone & \colorbox{rred}{\underline{0.19}} & \colorbox{rred}{\underline{0.13}} & \colorbox{oorange}{1.08} & \colorbox{oorange}{0.62} & \colorbox{oorange}{1.69} & \colorbox{oorange}{20.23} \\
MatAnyone 2 & \colorbox{rred}{\underline{0.19}} & \colorbox{rred}{\underline{0.13}} & \colorbox{rred}{\underline{0.99}} & \colorbox{rred}{\underline{0.46}} & \colorbox{rred}{\underline{1.13}} & \colorbox{rred}{\underline{15.19}} \\
\hline
\end{tabular}
}
\end{center}
\vspace{-7mm}
\end{table*}

\subsection{More Qualitative Comparisons}
\begin{figure}[h]
\begin{center}
    \includegraphics[width=\linewidth]{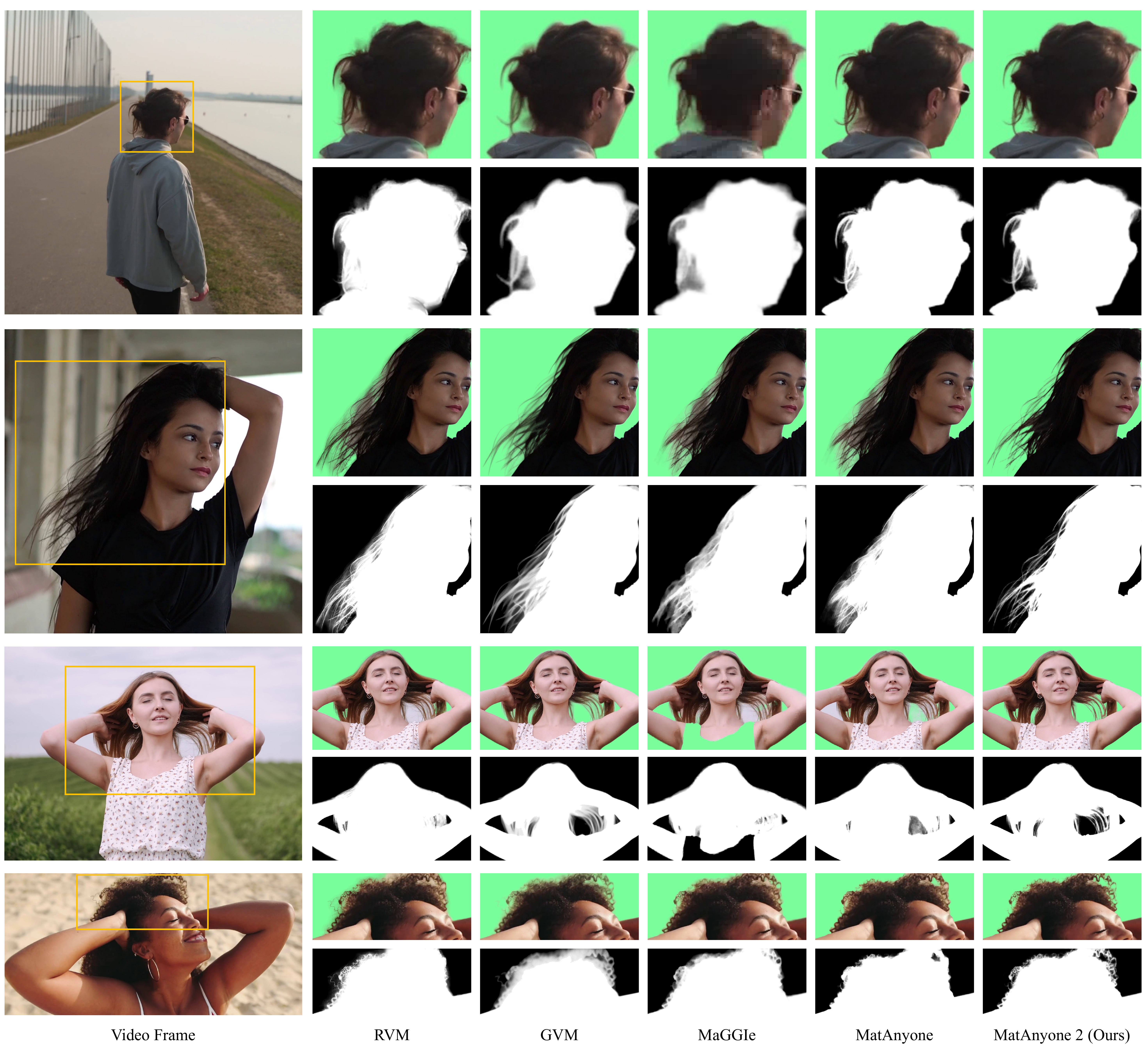}
    \vspace{-2mm}
    \caption{
    More qualitative comparisons on real-world videos.
    Our MatAnyone 2 significantly outperforms leading auxiliary-free (RVM~\cite{lin2022rvm}), diffusion-based (GVM~\cite{ge2025gvm}), and mask-guided (MaGGIe~\cite{huynh2024maggie} and MatAnyone~\cite{yang2025matanyone}) approaches in both detail extraction and semantic accuracy, especially under challenging real-world conditions.
     \textbf{(Zoom in for best view)}
    }
    \vspace{-4mm}
\label{fig:more_qualitative}
\end{center}
\end{figure}

\begin{figure}[h]
\begin{center}
    \includegraphics[width=\linewidth]{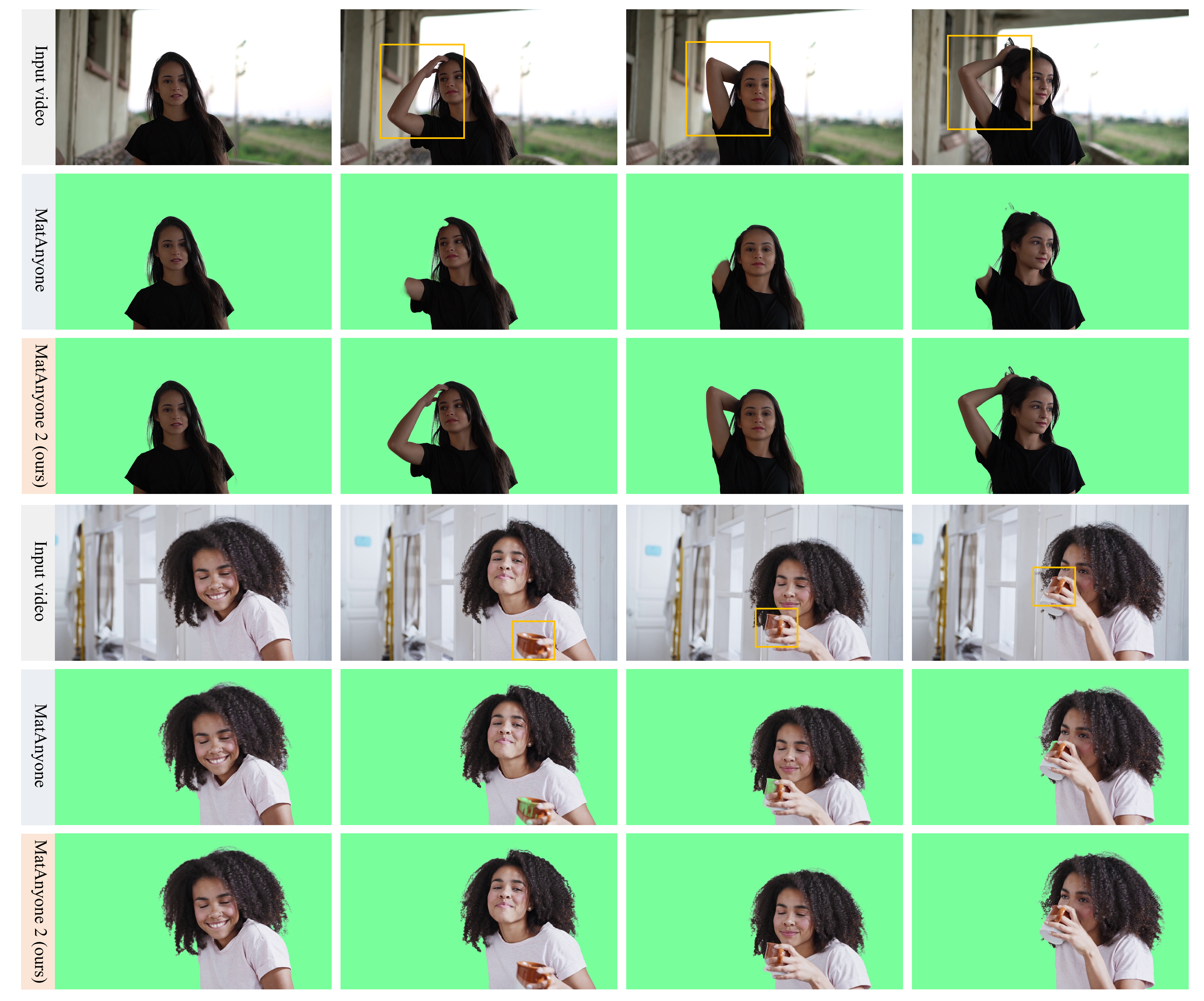}
    \vspace{-2mm}
    \caption{
    Robustness on long video sequences.
    This figure showcases the robustness of our method when previously unseen human body parts (\eg, arms) or handheld objects (\eg, a mug) appear in long video sequences. While MatAnyone fails to recognize these newly introduced regions, our method successfully recovers the complete structure with more accurate semantics and finer boundary details. 
    \textbf{(Zoom in for best view)}
    }
    \vspace{-4mm}
\label{fig:more_long}
\end{center}
\end{figure}

We provide additional visual comparisons of our method with several state-of-the-art methods, including auxiliary-free method RVM~\cite{lin2022rvm}, diffusion-based method GVM~\cite{ge2025gvm}, and mask-guided methods MaGGIe~\cite{huynh2024maggie} and MatAnyone~\cite{yang2025matanyone}. 
Fig.~\ref{fig:more_qualitative} further demonstrates the superiority of our model in various real-world challenging cases.
Moreover, Fig.~\ref{fig:more_long} showcases the robustness of our method on long video sequences, particularly when encountering previously unseen human body parts or handheld objects.

\subsection{Demo Video}
\label{subsec:demo_video}
We also offer a demo video in our \href{https://pq-yang.github.io/projects/MatAnyone2/}{[project page]}. This video showcases more video matting results on real-world challenging cases.

\section{Limitations}
\label{sec:limitations}

In this study, we introduce a scalable data annotation pipeline (with MQE) that combines the complementary strengths of state-of-the-art image and video matting models to construct VMReal, a large-scale, high-quality real-world video matting dataset. While the pipeline adaptively integrates semantic stability and fine boundary details from leading models, it remains bounded by their performance. A promising extension is to upgrade the pipeline into an iterative refinement process, where improved matting models progressively refine the alpha annotations, and in turn benefit from the enhanced data.  Such a closed-loop data-model ``refinement flywheel” could further boost both dataset quality and model performance, but would require considerable engineering effort and computational resources, and is thus left for future work. Nevertheless, the current VMReal dataset already offers strong supervision and yields notable performance gains across diverse video matting models.


\end{CJK}
\end{document}